\documentclass{ecai} 

\usepackage{latexsym}
\usepackage{amssymb}
\usepackage{amsmath}
\usepackage{amsthm}
\usepackage{booktabs}
\usepackage{enumitem}
\usepackage{graphicx}
\usepackage{color}

\usepackage{tabularx}
\usepackage{multirow}
\usepackage{xcolor}
\usepackage{url}
\usepackage[hidelinks]{hyperref}
\newcommand{\AR}[1]{\textcolor{black}{#1}}

\newcommand{\ar}[1]{\textcolor{black}{#1}}
\newcommand{\BP}[1]{\textcolor{black}{#1}}

\definecolor{darkgreen}{rgb}{0,0.5,0}

\newcommand{\Dom}{\ensuremath{\mathcal{D}}}

\newcommand{\model}{\ensuremath{f}}
\newcommand{\inn}{\ensuremath{\mathbf{x}}}
\newcommand{\simpinn}{\ensuremath{\mathbf{z}}}
\newcommand{\expl}{\ensuremath{e}}
\newcommand{\shap}{\ensuremath{\expl_{S}}}
\newcommand{\cfx}{\ensuremath{\expl_{O}}}

\newtheorem{definition}{Definition}
\newtheorem{example}{Example}

\newcommand{\BibTeX}{B\kern-.05em{\sc i\kern-.025em b}\kern-.08em\TeX}

\begin{document}


\begin{frontmatter}


\paperid{123} 


\title{Exploring the Effect of Explanation Content and Format on User Comprehension and Trust in Healthcare}


\author[A,B]{\fnms{Antonio}~\snm{Rago}
\footnote{Equal Contribution. Corresponding Authors. Emails: antonio.rago@kcl.ac.uk, b.palfi@gold.ac.uk.}}
\author[C]{\fnms{Bence}~\snm{Palfi}\footnotemark}
\author[A]{\fnms{Purin}~\snm{Sukpanichnant}}
\author[A]{\fnms{Kavyesh}~\snm{Vivek}}
\author[D]{\fnms{Hannibal}~\snm{Nabli}}
\author[A]{\fnms{Olga}~\snm{Kostopoulou}}
\author[A]{\fnms{James}~\snm{Kinross}}
\author[A]{\fnms{Francesca}~\snm{Toni}
}
\address[A]{Imperial College London, UK}
\address[B]{King's College London, UK}
\address[C]{Goldsmiths, University of London, UK}
\address[D]{University of Bristol, UK}


\begin{abstract}
 AI-driven tools for healthcare are widely acknowledged as potentially beneficial to 
    health practitioners and patients, e.g.~the QCancer regression tool for cancer risk prediction.
    However, for these tools to be trusted
    , they need to be supplemented with explanations
    .
    We examine how explanations' \emph{content} and \emph{format} affect user \emph{comprehension} and \emph{trust} when explaining QCancer's predictions. Regarding content, we deploy the SHAP 
    and Occlusion-1 explanation methods
    . Regarding format, we present SHAP explanations, conventionally, as charts (SC) and Occlusion-1 explanations as charts (OC) as well as text (OT), to which their simpler nature lends itself. 
    We conduct experiments with two sets of stakeholders: the general public (representing patients) and 
    medical students (representing healthcare practitioners).
    Our experiments 
    showed higher subjective comprehension and trust for Occlusion-1 over SHAP explanations based on content. However, when controlling for format, only OT outperformed SC
    , suggesting this trend is driven by preferences for text
    . Other findings corroborated that explanation format, rather than content, is often the critical factor.
    \end{abstract}

\end{frontmatter}


\section{Introduction}
\label{sec:intro}

The deployment of AI-driven tools in the healthcare domain has been widespread (see \cite{Tjoa_21} for an overview), in areas from image classification for brain tumour detection \cite{Legastelois_23} to clinical decision support \cite{Dominquez_23}. 
However, to foster trust in users, the underlying ``black-box'' AI models need to be supplemented with explanations with consideration of the types of stakeholders the users represent \cite{Langer_21}.
Explainable AI (XAI) has 
risen to prominence in recent years as a means of 
providing users with the explanations they need to be able to trust the 
outputs 
of AI models.
The ways in which explainability may facilitate trust in users has been studied extensively \cite{
Ferrario_22,
Ma_23}, with the general consensus leaning towards it playing an important role in this 
bond between humans and machines.
However, 
what remains to be seen is which forms of XAI 
are best suited 
in individual applications
of AI models \cite{Hamon_21}.
Further, given that the explanatory requirements 
within healthcare are different from those in general \cite{Legastelois_23}, such \emph{tailored} deployments are especially important in this domain \cite{Corti_24}.

One \ar{of the most} prominent \ar{types of} 
XAI method 
is \emph{feature attribution} (FA, see \cite{Bodria_23} for an overview). 
FA explanations assign each feature for a given input an attribution value denoting its importance towards the output. 
However, 
FA methods alone do not  account for the human factors involved in enabling trustworthy AI.
To evaluate how well XAI methods perform wrt these factors are human-centric metrics
such as \emph{(perceived) comprehensibility} \cite{Schmude_23}, i.e. how well users (think they, respectively) understand explanations, and \emph{trust} \cite{Jacovi_21}, i.e. how much users trust explanations.
However, 
only a limited number of studies, e.g. \cite{Szymanski_21,Bertrand_23,Schmude_23}, have considered the effect of the \emph{format}, rather than the \emph{content}, of explanations on these metrics.

One recent study \cite{Kostopoulou_22} investigates whether predictions from a cancer risk prediction model, namely the \emph{QCancer} algorithm \cite{hippisley2015development}, have a beneficial effect in supporting the referral decisions of UK-based General Practitioners (GPs). 
The 
study 
found that the inclination of the GPs on whether to refer the patient changed 26\% of the time, while their decisions switched entirely only 3\% of the time. However, despite the reluctance of 
 GPs to 
 trust and adopt advice from the algorithm
, the changed decisions were 
more consistent with established clinical guidelines, 
highlighting the potential of this approach to decision support.
In a follow-up study, P\'alfi et al.~\cite{palfi2023risk} found that providing a naive, chart-based FA explanation that visualised the relative weights (based on the regression coefficients of the model) of the features for each prediction did not improve the trust in and uptake of the algorithm, speculating that this could be related to the GPs' difficulty in comprehending the explanations.




In this paper, we set out to address some of the unanswered questions 
highlighted by 
P\'alfi et al.~\cite{palfi2023risk}, broadening from healthcare practitioners to include potential patients also.
Namely, we 
undertook a study involving different stakeholders, 
aiming to shed light on differing degrees of (perceived) comprehensibility and trust affected in users when they receive explanations of different 
contents and formats.
\ar{We restrict our focus to FA explanations as in \cite{palfi2023risk}, and leave the study of whether our findings translate to other XAI methods to future work.}
We 
\ar{undertake this study}, in collaboration with healthcare practitioners, by deploying different, well-known FA 
explanation methods, \emph{SHAP}~\cite{Lundberg_17} and \emph{Occlusion-1}~\cite{Ancona_18} \ar{(two of the most cited works in this area)}, for the QCancer algorithm 
\ar{in an online system, providing a useful tool for future research}.
We undertook two user studies, each with 
a different group of participants, one sampled from the general population and one from students in medical training, to examine the effect of different 
stakeholders' expertise and explanation goals \cite{Delaney_23}, which 
are important factors in the evaluation of explanations \cite{Kim_23,Ehsan_24}. 
The participants were asked to 
take part in a survey to assess their (perceived)
comprehensibility of and trust in explanations for the outputs of the QCancer algorithm on fictitious patients described by vignettes from \cite{palfi2023risk}. Moreover, we created a novel objective measure of comprehension (differentiating from perceived comprehension) to control for any potential reporting biases.
Differently from existing studies, we control for the format of explanations (\emph{chart} or \emph{text}) to examine their effect on this paradigm.
We thus aim to explore the effect of explanation content and format on
users' comprehension and trust via the following research questions (RQs), all considered within the healthcare context:\\
%
%
    \noindent
    \textbf{RQ1:} 
    Are \AR{Occlusion-1} explanations, in \textbf{a}.)~chart or \textbf{b}.)~text format, \emph{more comprehensible} than \AR{SHAP} explanations? \\
    \noindent
    \textbf{RQ2:} 
    Are \AR{Occlusion-1} explanations, in \textbf{a}.)~chart or \textbf{b}.)~text format, \emph{perceived as being 
    more comprehensible} than \AR{SHAP} explanations
    ? \\
    \noindent
    \textbf{RQ3:} 
    Are \AR{Occlusion-1} explanations, in \textbf{a}.)~chart or \textbf{b}.)~text format, \emph{perceived as being more trustworthy} than \AR{SHAP} explanations
    ?

Our expectations were for the null hypothesis to be rejected 
(i.e. the RQs would be answered affirmatively), given the 
complexity in SHAP explanations. We were surprised to discover instead that the dominance of the simpler Occlusion-1 explanations 
is unclear and, where present, 
may be due to the use of the textual format. These findings 
provide 
guidance for the future design of AI-based decision support in healthcare.



\section{Related Work}
\label{sec:related}





\paragraph{User Studies Comparing 
Types of XAI}
Some works, e.g. 
\cite{Bertrand_23}, 
also assess FA explanations in text and chart formats
, finding that the former increase trust relative to the latter, possibly due to their simplicity. 
%
Others have reached similar conclusions, e.g. Szymanski et al.~\cite{Szymanski_21} assess visual and textual explanations wrt comprehensibility, controlling for user expertise, as we do, but find that lay users prefer visual explanations while often misinterpreting them compared to textual ones. 
%
Finally, Schmude et al.~\cite{Schmude_23} evaluate user comprehension of textual, dialogical 
and interactive 
explanations, 
finding that 
the former result in worse comprehension and lower preferences 
from users. 

Many works 
show the importance of 
expertise with regard to user comprehension and trust, e.g. \cite{Wang_23
}, and call for explanations to be designed to suit the needs of individuals \cite{Lesley_24} and 
make sense to them \cite{Byrne_23}.
Other works 
validate our design choices, e.g. our decision to differentiate between self-reported and objective comprehensibility \cite{Cheng_19}
.
Our approach is a step towards the effective design of multi-modal explanations, e.g. as in 
\cite{Szymanski_23} in the healthcare domain.

None of these works 
assess (perceived) comprehensibility and trust of FA 
explanations, controlling for the 
format and the participant expertise, as we do.





\paragraph{User Studies of XAI in Healthcare}
XAI is often considered within healthcare as a tool to enhance trust in and uptake of prediction models \cite{Tjoa_21
}. 
Lebovitz et al.~\cite{Lebovitz_22} conducted a user study in which AI 
was deployed 
by radiologists 
to predict breast cancer, lung cancer and bone age: they
found that trust in the AI models' predictions, and thus incorporation into the radiologists' judgements, were highly dependent on their department, pointing to customised XAI as a potential solution.
Moreover, a recent vignette-based user study found promising results \cite{Naiseh_23} in XAI's deployment in healthcare, finding the AI 
to be more comprehensible when they received an explanation compared to when they did not. 

Some studies obtained 
less promising results. For instance, in a user study with healthcare providers, Panigutti et al.~\cite{Panigutti_22} assessed the effect of explanations on trust in AI 
operating a clinical decision support system. 
They showed that users preferred explanations from the AI model, though the user satisfaction with the explanations themselves was low. 
Wysocki et al.~\cite{Wysocki_23} found that when receiving explanations for COVID-19 risk predictions, users found FA charts easy to interpret but there was no evidence that providing them 
increased 
trust in and satisfaction with the AI
, compared to when receiving predictions without explanations.
A vignette-based study with intensive care doctors found no 
benefit of an FA explanation over a prediction only option for prescription decisions \cite{nagendran2023quantifying}. 
Finally, a mixed-methods, vignette-based study found that chart-based FA and text-based counterfactual explanations failed to meet the expectations of physicians in reducing uncertainty around heart disease diagnosis 
\cite{Lesley_24}.

In summary, 
few works 
test XAI's benefits for healthcare with user studies, 
with mixed results, which is a known deficiency in the field~\cite{
Sadeghi_24}. 



\section{Methodology}
\label{sec:methods}



\paragraph{Vignettes}
We adapted twenty fictitious vignettes (including one practice vignette) from \cite{palfi2023risk}, describing hypothetical patients presenting at the primary care (i.e. their GP) with symptoms and risk factors suggestive of gastro-oesophageal cancer. 
We chose to focus on one type of cancer here for simplicity, though the QCancer algorithm and the original vignettes concern multiple types
.
Each vignette beg
ins with a table of patient demographics and some risk factors, and continue
s with a textual description of the patient's symptoms, other risk factors and the results of some relevant examinations. All of the presented factors were predictive of gastro-oesophageal cancer according to the QCancer algorithm. Then, we provided the estimate of the patient's risk of gastro-oesophageal cancer calculated by the QCancer algorithm with an explanation of this prediction and information on how the explanation should be interpreted.
An example of a vignette, with the output of the QCancer algorithm shown for gastro-oesophageal cancer, is given in the following. 

\begin{example}
\label{ex:FA}

    Patient name: Joyce Acorn (female),
    Age: 65,
    BMI: 30.64 (177cm, 96kg),
    Smoking: 
    20 cigarettes per day,
    Past medical history: Nil, 
    Family history: Nil. 

    Joyce comes to see you about the indigestion she's been experiencing over the past month. She also mentions that it has become increasingly difficult to swallow during meals. She enjoys food usually but has noticed that she is now off her food. She denies any post-menopausal bleeding. When she saw the practice nurse last week for her flu vaccine, she had also taken some routine blood tests. There are no other symptoms and examination findings are normal.

    The cancer prediction algorithm estimates the risk of gastro-oesophageal cancer to be 31.90\%. 
\end{example}

The 
variables used in the QCancer algorithm's prediction of gastro-oesophageal cancer 
and the values in Example \ref{ex:FA} are shown in Table \ref{table:qcancer}\AR{, along with the}  
the baseline patient values for males and females, which will be discussed next.

\begin{table*}[ht]
\centering
\begin{tabular}{cccccccccccccccc}
\cline{2-16}
 &
Gender & 
Age & 
BMI &
SS &
T2 &
CP &
LA &
WL &
DS &
AP &
BV &
I &
H &
A & 
BH \\ 
\hline
Baseline Male &
M &
40 &
25 &
non &
no &
no &
no &
no &
no &
no &
no &
no &
no &
no &
no \\
Baseline Female &
F &
40 &
25 &
non &
no &
no &
no &
no &
no &
no &
no &
no &
no &
no &
no \\
Example \ref{ex:FA} &
F &
65 &
30.64 &
heavy &
no &
no &
yes &
no &
yes &
no &
no &
yes &
no &
no &
no \\
\hline
\end{tabular}
\caption{Values for the QCancer algorithm used for the baseline patients and the patient in Example \ref{ex:FA}, with variables including 
Smoking Status (SS), Type 2 Diabetes (T2), Chronic Pancreatitis (CP), Loss of Appetite (LA), Unintentional Weight Loss (WL), Difficulty Swallowing (DS), Abdominal Pain (AP), Blood in Vomit (BV), Indigestion (I), Heartburn (H), Anaemia (A) and Change in Bowel Habit (BH). 
}
\label{table:qcancer}
\end{table*}

The vignettes were divided into a low 
and high 
risk sets based on the estimated risk of gastro-oesophageal cancer for the patient computed by the QCancer algorithm\BP{ and rounded to the closest integer. We based the threshold for high-risk cases on the 3\% referral threshold for suspected cancer of the NICE guidelines for UK healthcare professionals\footnote{\url{https://www.nice.org.uk/guidance/ng12/evidence/full-guideline-pdf-2676000277}}.} 
To control for order effects and to avoid the repetition of vignettes during the experiment, the sets of low and high risk vignettes were further divided into two subsets of vignettes with similar risk profiles, giving low risk sets A and B, and high risk sets A and B 
(more details are given later in this section).

\paragraph{Explanations for the QCancer Algorithm}
To define the explanations that we deliver to users we assume a simple, single-label classification problem comprising a classifier $\model: \Dom_1 \times \ldots \times \Dom_k 
\rightarrow [0,1]$, where $\Dom_1, \ldots, \Dom_k$ represent the domains for each of the $k$ features ($k=15$ in our setting, see Table~\ref{table:qcancer}). 
Then, for any input $\inn \in \Dom_1 \times \ldots \times \Dom_k$, $\model(\inn) \in [0,1]$ is the predicted probability of the class given the current values of the input features.
This represents the QCancer algorithm, where the features, e.g. \emph{age}, are inputs to the algorithm  and the output is the predicted risk of gastro-oesophageal cancer.
We refer to the $i$th value of any input $\inn$ as $x_i$, where $i \in \{ 1, \ldots, k \} $.
In order to define the FA  explanations, we first introduce a simplified version of any given input $\inn$ as $\inn'$, where there exists a mapping function $h_\inn$ such that $\inn = h_\inn(\inn')$, i.e. returning the original input given the simplified version.
We then let, as in \cite{Lundberg_17}, $\inn' \in \{ 0, 1 \}^M$, where $M$ is the number of simplified input features (note that $M = k$ throughout this paper). Then, if $x_i' = 1$, then $\inn = h_\inn(\inn')$ sets $x_i$ to the current value of the feature, while if $x_i' = 0$, then $\inn = h_\inn(\inn')$ sets $x_i$ to the feature's \emph{baseline value}, i.e. a pre-defined default value.
We set the baseline values (referred to as \emph{baseline patients} in the explanations) to 
one of two sets of values representing healthy males and females depending on the patient's gender, as indicated in Table \ref{table:qcancer}
, which were used for the 
studies.


One of the most prominent of all 
FA methods 
is 
SHAP \cite{Lundberg_17}, 
where each feature of an input to a model is assigned 
an importance value computed from a game-theoretical approximation of its contribution. 

\begin{definition}
\label{def:fa}
    Given a classifier $\model$ and an input $\inn$, the SHAP explanation for $x_i$ wrt $\model$ is:
    \begin{align}
        \shap(\model,\inn,i) \!=\!\! \sum_{\simpinn' \subseteq \inn'}{\frac{|\simpinn'|!(M \!-\! |\simpinn'| \!-\! 1)! }{M!} \model( h_\inn (\simpinn')) \!-\! \model(h_\inn (\simpinn_{-i}' 
        ))} \nonumber
    \end{align}
    where $\simpinn' \subseteq \inn'$ means 
    that $\simpinn'$
    is a vector whose non-zero entries are a subset of those in $\inn'$, 
    $|\simpinn'|$ is the number of non-zero entries in $\simpinn'$
    and $\simpinn_{-i}' 
    $ denotes $\simpinn'$ after setting $z_i' = 0$. 
\end{definition}

For a given input, SHAP assigns each feature an attribution value based on a weighted sum considering the effect of changing the feature from its current value to the baseline value in a range of combinations of other features' values.

\begin{figure}[t]
    \centering
    \includegraphics[width=0.9\linewidth]{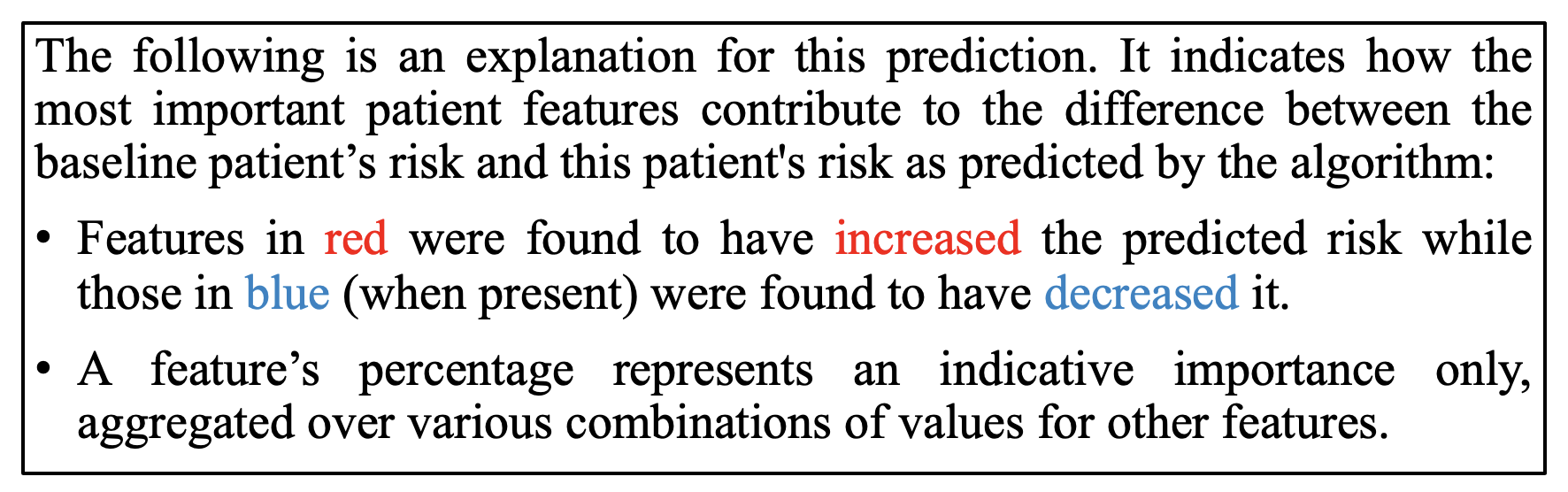}
    \includegraphics[width=\linewidth]{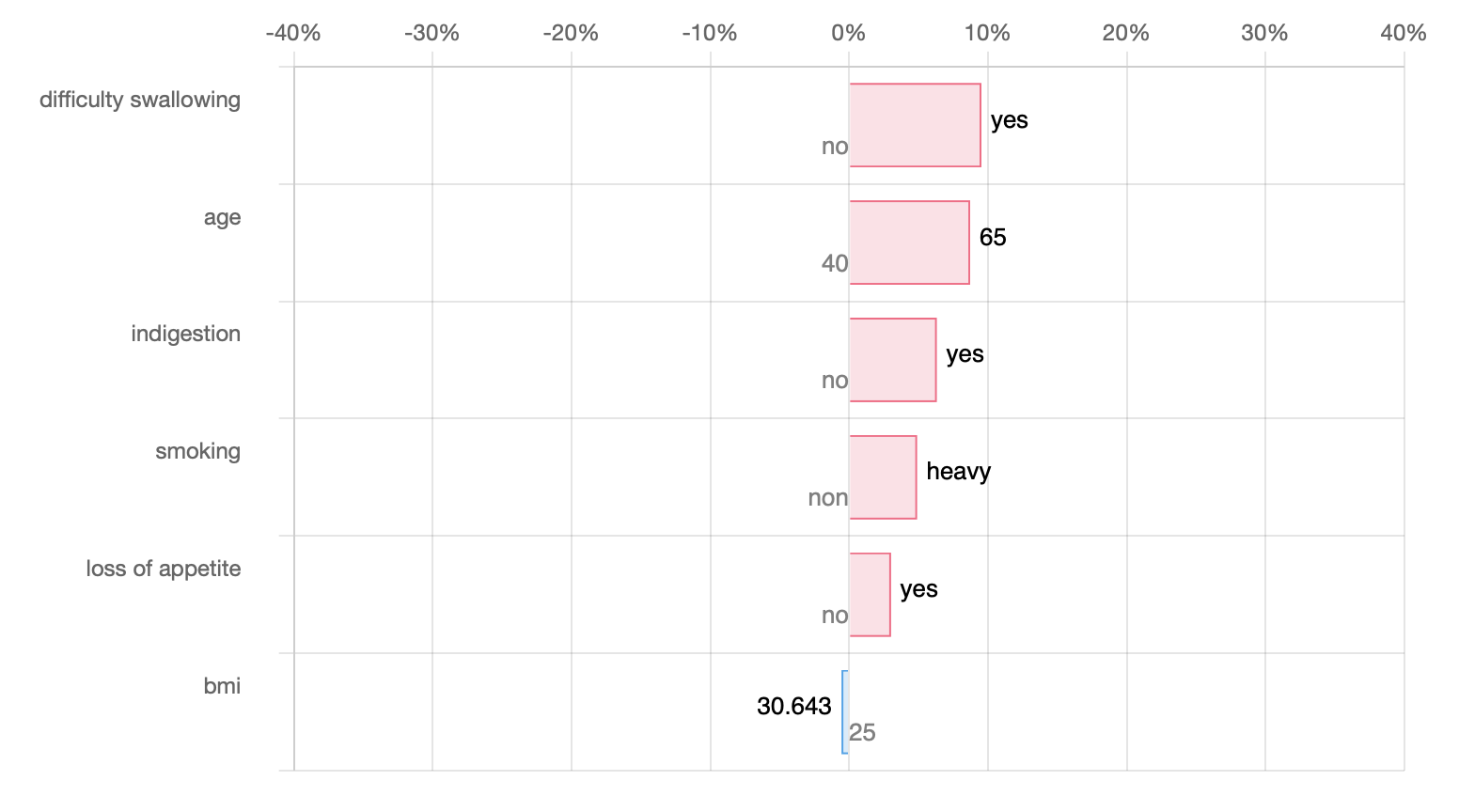}
    \protect\caption{\AR{SC} explanation (bottom) for the vignette given in Example \ref{ex:FA}, and textual description thereof (top). Red (blue) bars represent features which increased (decreased, respectively) 
    the risk of cancer. The 
     patient's values are given in bold and the baseline values 
     in grey
     .}
    \label{fig:acornFA}
\end{figure}

We then introduce \emph{chart-based \AR{SHAP} (\AR{SC}) explanations}, which are the 
explanations computed by SHAP expressed as a percentage in chart format, as exemplified in Figure \ref{fig:acornFA} (bottom). 
For all explanations, 
the displayed features were the six with the highest magnitudes (ordered as such).
Note that all explanations were accompanied by a textual description of the explanation, as 
shown in Figure \ref{fig:acornFA} (\AR{top}).
In the example explanation in Figure~\ref{fig:acornFA}, 
we can see that the top five features increased the risk of gastro-oesophageal cancer, with \emph{difficulty swallowing} and its value of \emph{yes} being the most important. The last feature, \emph{BMI}, with value \emph{30.643}, reduced the risk but had a smaller magnitude, and thus importance, as calculated by SHAP.

\AR{We then selected the Occlusion-1 method \cite{Ancona_18} for its simplicity in hopes of achieving more comprehensibility and trust in users.}
\AR{This method takes} the set of baseline feature values (as in SHAP explanations) and assigns any feature an attribution corresponding to the difference in the prediction if the feature's value is changed to the baseline value (with all other features' values remaining unchanged). 

\begin{definition}
\label{def:cf}
    Given a classifier $\model$ and an input $\inn$, the \emph{
    \AR{Occlusion-1} explanation} for $x_i$ wrt $\model$ is:
    \begin{align}
        \cfx(\model,\inn,i) = \model(h_\inn (\inn')) - \model(h_\inn (\inn_{-i}' 
        )) \nonumber
    \end{align}
    where $\inn_{-i}' 
    $ denotes $\inn'$ after setting $x_i' = 0$. 
\end{definition}

Note that, while 
\AR{these} explanations do not explicitly mention datapoints, each feature's attribution in explanations implicitly refers to a modified datapoint with that feature set to the baseline value, with all other features remaining unchanged.
Thus, while this differs from counterfactual and semi-factual explanations in the literature (see \cite{Guidotti_22,Aryal_23}, respectively, for recent overviews
), which are based on single modified datapoints, it still gives an actionable representation of (six) modifications users could make to the input to achieve a different output \AR{(as in Albini et al.~\cite{Albini_20})}. 
\AR{We thus tentatively predicted that this inherent counterfactual nature of Occlusion-1 explanations may give more amenability to humans, as observed for counterfactual explanations by Miller \cite{Miller_19}, which strengthened our prediction of the rejection of the null hypotheses for \textbf{RQ1-3}.}

\begin{figure}[t]
    \centering
    \includegraphics[width=0.9\linewidth]{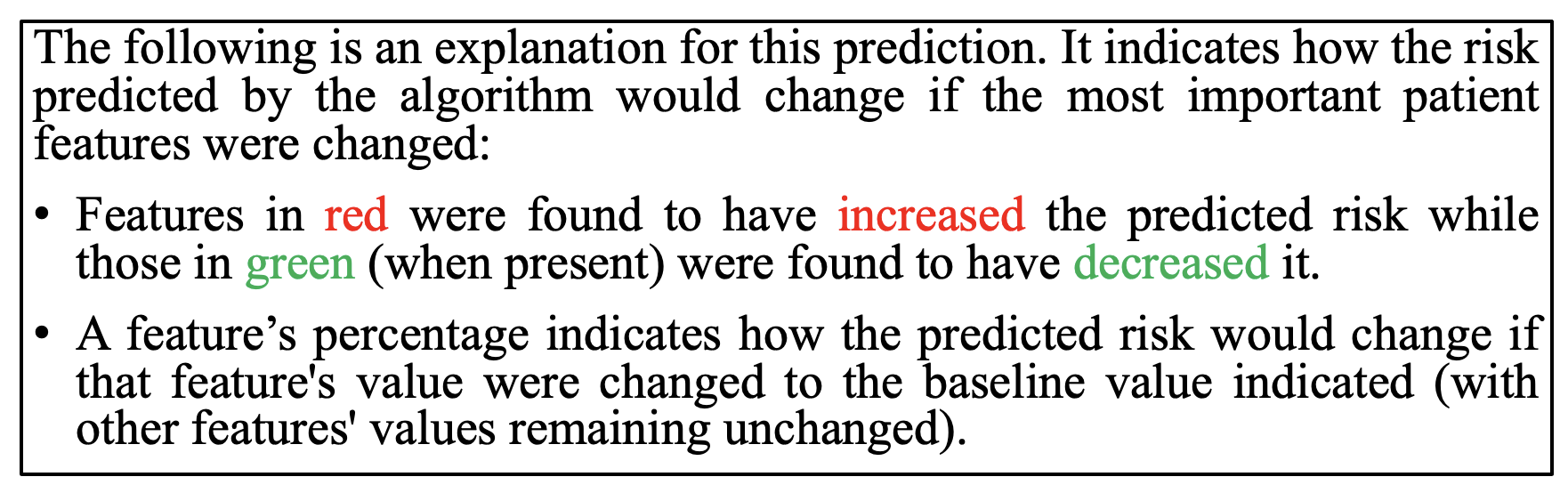}
    \includegraphics[width=\linewidth]{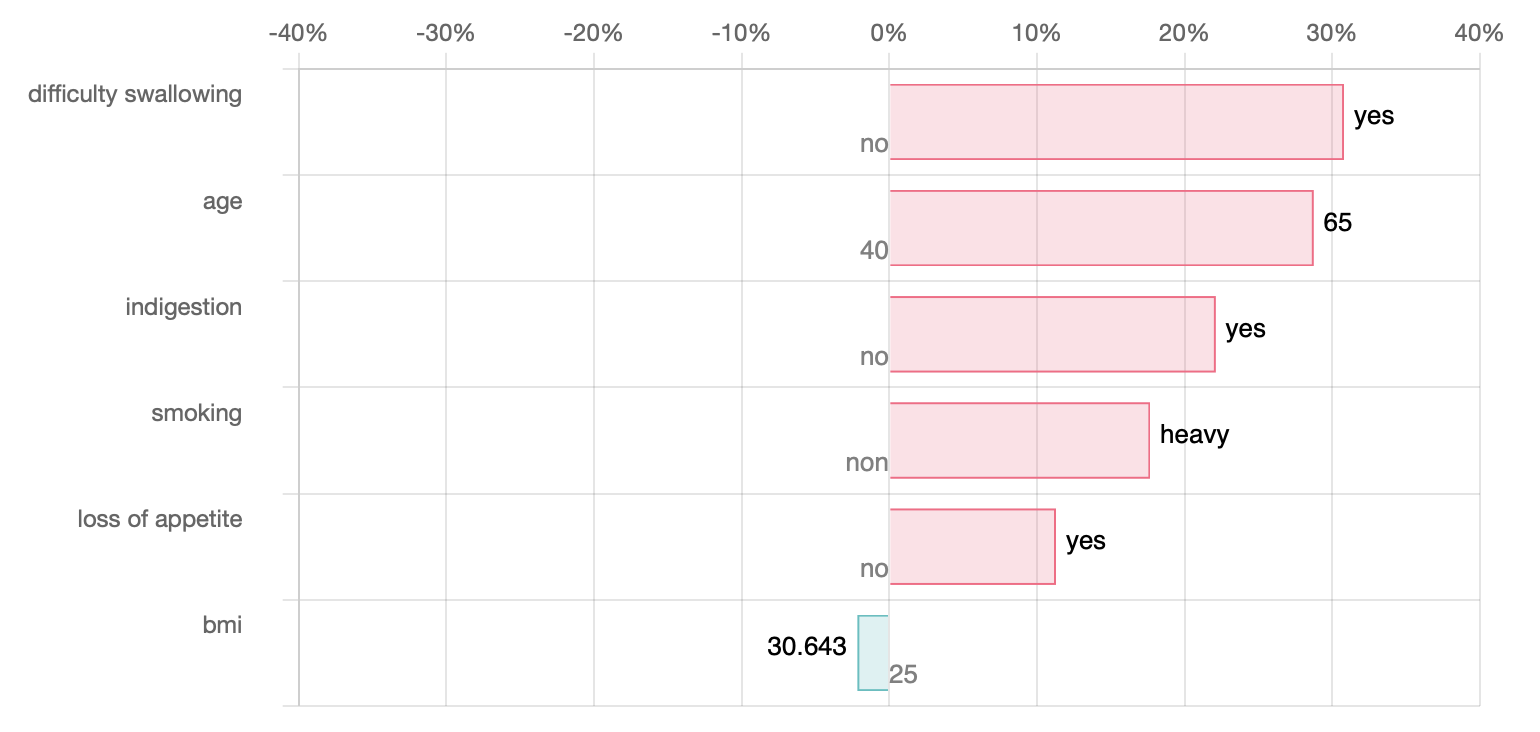}
    \protect\caption{\AR{OC} explanation (bottom) for the vignette given in Example \ref{ex:FA}, and textual description thereof 
    (top). Red (green) bars represent features which increased (decreased, respectively) 
    the risk of cancer. The 
    patient's values are given in bold and the baseline values 
    in 
    grey
    .}
    \label{fig:acornCC}
\end{figure}

\begin{figure}[t]
    \centering
    \includegraphics[width=0.9\linewidth]{images/OccExp.png}
    \includegraphics[width=0.9\linewidth]{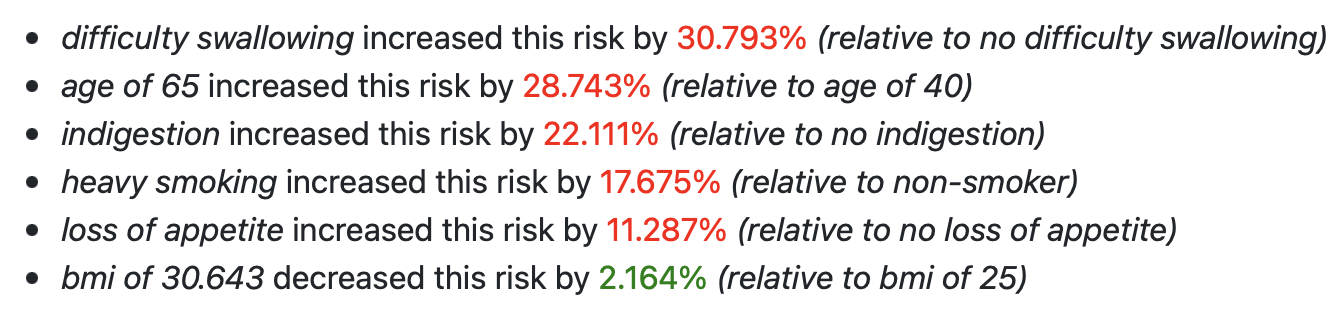}
    \protect\caption{OT explanation \AR{(bottom)} for the vignette given in Example \ref{ex:FA}, and textual description thereof 
    (top). Note that the textual descriptions are identical for the \AR{OC} and \AR{OT} explanations.  
    }
    \label{fig:acornCT}
\end{figure}

This definition \AR{thus} allow\AR{ed} us to introduce two different formats of 
explanation. Firstly, \emph{chart-based 
\AR{Occlusion-1 (OC)} explanations} are the 
explanations computed as in Definition \ref{def:cf} in chart format as in the SC explanations, as shown in Figure~\ref{fig:acornCC} with 
an accompanying textual description. 
We can see here that the features are the same as those selected by the 
\AR{SC} explanations but the values are significantly different
.
\AR{Further, the counterfactual reading of these explanations also allowed us to describe the attributions intuitively to users.}
Thus, we used the same 
\AR{Occlusion-1} explanations, this time in textual format, to define \emph{text-based \AR{Occlusion-1 (OT)} explanations}, which are delivered as shown in Figure \ref{fig:acornCT} with their accompanying textual description.
This distinction between the two types of 
\AR{Occlusion-1} explanation allows us to assess the effects not only of the content of an explanation, but also 
of its format.\footnote{SHAP allowed no such intuitive reading: 
the complexity of SHAP in Definition \ref{def:fa} demonstrates they how do not lend themselves to text in what we might have called ``ST'' explanations, and that most users would not understand these game-theoretical notions. Indeed, it is difficult to picture, at least in a cognitively manageable form, what the ST equivalent for the SC explanation in Figure \ref{fig:acornFA} would be, as the OT explanation in Figure \ref{fig:acornCT} is to the OC explanation in Figure \ref{fig:acornCC}. We could have given an approximation of these explanations, but this simplification could be seen as being unrepresentative of SHAP, and it would be an unfair test. Thus, we refrained from defining and experimenting with this form of explanation.}


\begin{figure*}[t]
\centering
    \includegraphics[width=0.9\linewidth]{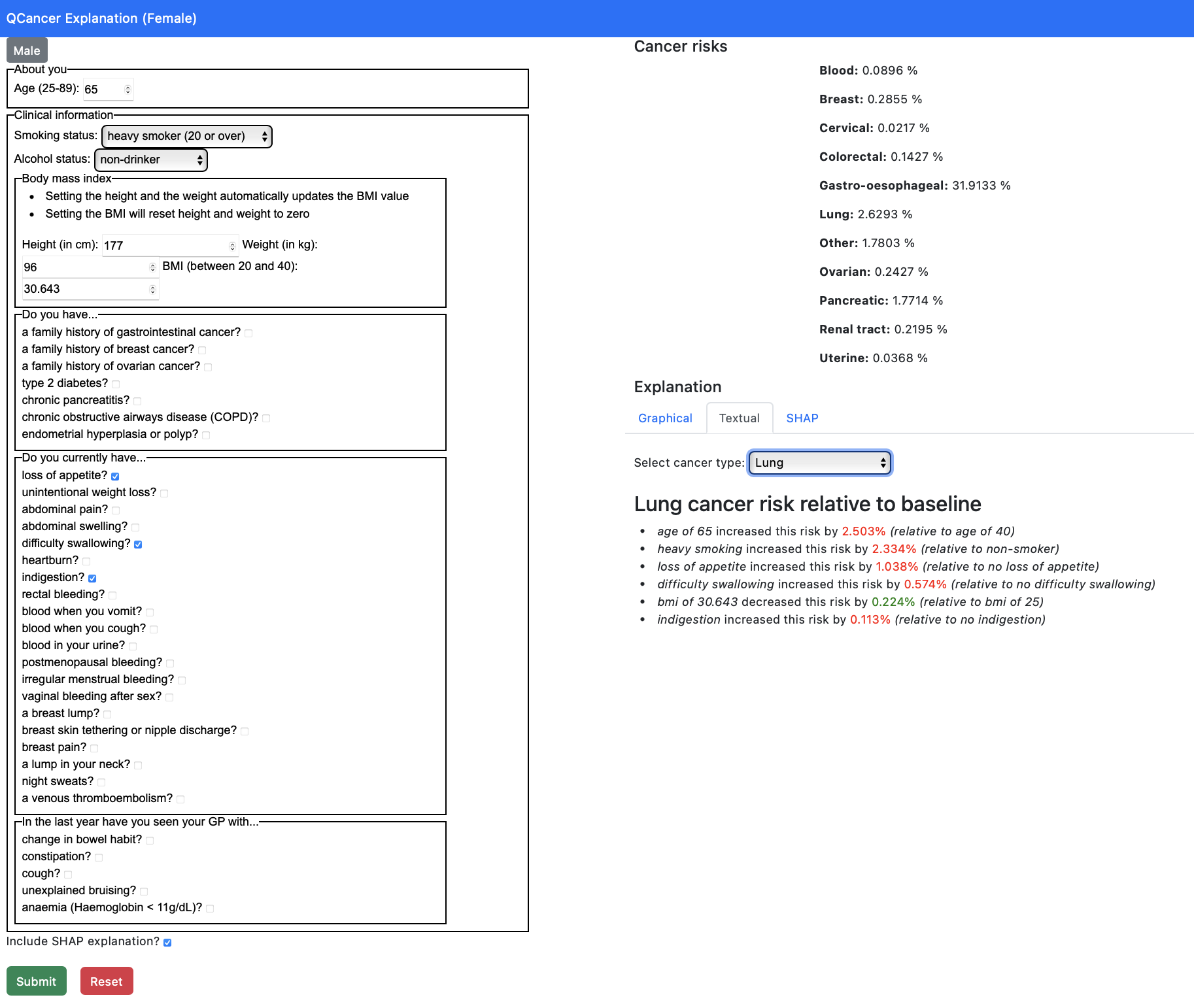}
    \protect\caption{
    \ar{User interface for our online tool for visualising different forms of explanation for the QCancer algorithm, showing the OT explanation for the patient from Example \ref{ex:FA} and the next highest predicted type of cancer (lung) after gastro-oesophageal.
    }}
    \label{fig:system}
\end{figure*}

\ar{All explanations in Figures \ref{fig:acornFA} to \ref{fig:acornCT} were visualised with the purpose-built online tool\footnote{\url{https://qcancer-explanation.herokuapp.com/\#}} that we developed for explaining the prdictions of the QCancer algorithm, the user interface for which is shown in Figure \ref{fig:system}.
SHAP explanations remain optional in the system due to their additional computation time. It should be noted that more features are included here (compared with Table \ref{table:qcancer}) as the online tool can be used to explain QCancer predictions for all supported types of cancer, not only gastro-oesophageal cancer as is our focus in this paper. Our hope is for this tool to help support future (user) studies of XAI for the QCancer algorithm.
}

\paragraph{Experimental Design and Procedure}
We assessed the RQs in two different samples: 
one from the general population where a limited number of participants had medical experience, representing stakeholders in the form of patients (Study 1), and an expert sample where participants had 
some medical training and experience, representing stakeholders in the form of healthcare practitioners (Study 2). 
We created two Qualtrics\footnote{\url{https://www.qualtrics.com}
} surveys using the following structure (see the Supplementary Material 
for an example): 



\begin{itemize}
    \item Pre-Test (6 or 8 questions) -- The participants were shown information about the study and were asked to provide consent. They were then asked questions regarding their background and experience of AI and cancer prediction algorithms (
    additional demographics questions were included for Study 2), including their trust in AI models in healthcare, before the task was explained to them with an illustrative example. 
    \item Main Test (4 x 6 questions) -- In the main task, we provided the participants with four vignette sets, each containing a vignette, a prediction and an explanation (with its textual description), along with six questions on the explanation. The four sets concerned two types of explanation for comparison, each with a low risk and a high risk vignette. The six questions consisted of:
    \begin{itemize}
        \item an attention check 
        comprising the most important feature task (attention check, free text variable); 
        \item a perceived comprehensibility rating (addressing \textbf{RQ2}, interval scale from 0 to 10); 
        \item a trust rating (addressing \textbf{RQ3}, interval scale from 0 to 10); 
        \item a request for reasoning for the trust rating (optional, free text variable); 
        \item a request for additional feedback (optional; free text variable); 
        \item a comprehensibility test comprising the definition recognition task (addressing \textbf{RQ1}, categorical variable with four options).
    \end{itemize} 
    \item Post-Test (4 questions) -- The participants were asked to leave feedback if they had any technical difficulties during the survey, if they found the reimbursement fair and if they had any general comments about the study.
\end{itemize}

After each vignette, the participants were provided with an the attention check to assess whether they were engaging with the experiment. The attention check consisted in asking participants which was the most important feature 
in determining the gastro-oesophageal cancer risk according to the explanation, with the correct answer being the feature which gave the highest absolute value of $\shap$ for \AR{SC} explanations and $\cfx$ for \AR{OC} and \AR{OT} explanations.
For the perceived comprehensibility rating (\textbf{RQ2}) and the trust rating (\textbf{RQ3}), we asked the participants how well they thought they understood and how much they would trust, respectively, the explanation, on a scale of 0-10. 
Next, we asked for free-form textual feedback from the participants on reasons for their trust rating and on any additional points, both of which were optional.
Finally, on a new page, we 
measured the participants' comprehension of the explanation (\textbf{RQ1}) by offering four differently worded descriptions of the explanation as indicated below, one of which was correct for each explanation (as indicated in square brackets, but not shown to participants), and asked them which one they believed to be correct, as follows.

\emph{What would you say best describes a feature’s percentage indicated in the explanation on the previous page?}

    %
    \noindent
    $\bullet$ \emph{The change in predicted risk when the feature’s current value is modified to the baseline value.} [Correct for OC/OT.]
    %
    
    \noindent
    $\bullet$ \emph{A representative importance of this feature’s value relative to the baseline value, based on different configurations of values for other features.} [Correct for SC.]
    %
    
    \noindent
    $\bullet$ \emph{The portion of the predicted risk associated with the feature.}
    %

    \noindent
    $\bullet$ \emph{The average predicted risk for a patient with the feature’s current value.}
    %

\begin{figure}[t]
\centering
    \includegraphics[width=1\linewidth]{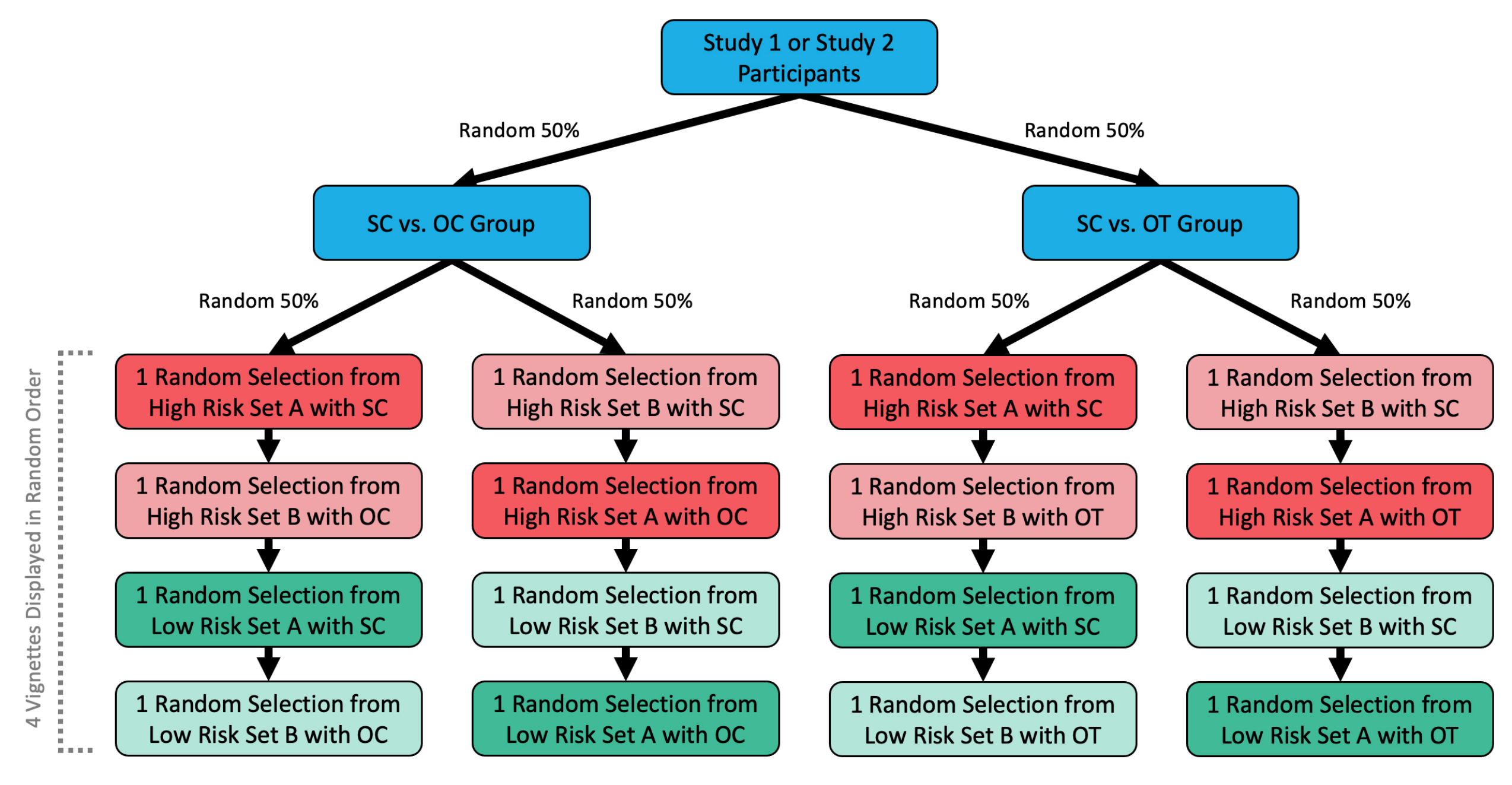}
    \protect\caption{Experimental design for both user studies, with the two sets of high risk vignettes represented by shades of red, and the two sets of low risk vignettes represented by 
    shades of green.}
    \label{fig:structure}
\end{figure}

Figure \ref{fig:structure} illustrates the experimental design of Studies 1 and 2, which were identical in this regard.
The participants were split such that 50\% assessed \AR{SC} versus \AR{OC} explanations (i.e. addressing \textbf{RQ1a}, \textbf{RQ2a} and \textbf{RQ3a}) and 50\% assessed \AR{SC} versus \AR{OT} explanations (i.e. addressing \textbf{RQ1b}, \textbf{RQ2b} and \textbf{RQ3b}). We then split the twenty vignettes into one vignette for demonstration to all participants, ten vignettes presenting low risk patients 
and nine vignettes presenting high risk patients 
(as defined earlier), to ensure that all participants received one low and one high risk for each of the explanations that they examined. We then further split the high and low risk sets into two (giving High Risk Sets A and B, and Low Risk Sets A and B) so that 
all vignettes were presented evenly. Further, the four vignette sets for each participant were presented in random order to ensure there was no ordering bias.

\paragraph{Participant Recruitment}
For Study 1, we aimed to recruit 100 participants from the general population, 
who were over 18 and spoke English via the online research platform Prolific\footnote{Prolific (\url{https://www.prolific.com}
) was chosen 
for its superior data quality when compared with other platforms \cite{Peer_22}.}. Upon completion, the participants were paid an average of \$4.50 for their time based on the platform’s standard reimbursement rates of \$21.20 per hour. Data collection was anonymous during Study 1. In Study 2, we aimed to recruit at least 50 students in medicine 
from 
UK-based universities. The potential participants were approached via student mailing lists, and were invited via a link to the survey in an email, which also contained the participant information sheet with details about the study.
All students were at least 18 years of age and English-speaking as they were attending university in the UK. 
At the end of Study 2, participants were asked to provide their university email address, to which we sent a £10 voucher for the online retailer Amazon to reimburse them
.\footnote{The anonymised 
data 
is available at \url{https://zenodo.org/records/16901820}.}

\ar{Participation in both studies was voluntary. Participants were asked to provide informed consent after reading 
an information sheet and they could withdraw at any time prior to 
completion, simply by closing the 
survey window, with 
their data 
deleted permanently. We informed the participants that once they had completed the survey, they could no longer withdraw and we would keep their data.
All the participant responses were 
anonymised once payment was made (i.e.~we deleted their e-mail addresses from the dataset). 
We received ethics approval from the institution leading these studies.}



\paragraph{Data Analysis}
We created categorical variables with two levels from the responses to the most important feature and the definition recognition tasks, where each response was categorised as either correct or incorrect. When more than one feature was named as the most important, we categorised the response as correct as long as the first mentioned feature was the one with the strongest absolute contribution. Moreover, for the vignettes where the feature with the highest absolute contribution was a protective factor (i.e. the reduced risk of cancer), we accepted responses as correct where the strongest negative contributing feature was named first. We used the subjective understanding of the explanations and the trustworthiness of the explanations variables as interval variables.

We ran the statistical analyses in R (version 4.3.1). We built generalised linear and linear mixed-effects regression models with the lme4 R package \cite{lme4} to analyse our categorical and interval variables, respectively. For the generalised linear regression models, we only included random intercepts by participants
, and for the linear regression models, we included random intercepts by participants and vignettes. For the generalised linear and the linear regression models we report Odds Ratios (\textit{OR}s) and raw regression slopes (\textit{b}s), respectively. We also report 95\% CIs for each regression coefficient. To test our hypotheses, we used 
a p-value of 0.05.

We ran two types of analyses for RQ2 and RQ3. First, we compared SHAP explanations to a single Occlusion-1 explanation group, merging OC and OT explanations, for a general test of explanation content. Second, we made pairwise comparisons of explanations to examine the effects of explanation content, format and their interaction. We did not conduct formal analyses of the free-text responses of the comments and justifications on trust towards the explanations due to the low response rates 
(15\% across questions and both studies).



\section{Results}
\label{sec:results}


\paragraph{Study Samples}
Study 1 was completed by 99 participants, from which 1 reported to have colour-blindness (we did not measure demographics). \BP{Since 5 participants failed the attention check, we did not include their data in any of our analyses (see more details below) resulting in a final sample of 94.} Study 2 was completed by 69 medical students, from which 3 reported to have colour-blindness.\footnote{One participant skipped the survey in Study 1 and one participant left a non-university email in Study 2; we deleted the data of both participants and did not include them in the results.} The mean age of the participants in Study 2 was 22.3 (\textit{SD} = 3.3), with 35 females (32 males, 1 other and 1 
preferred not to disclose). As expected, participants in the medical student sample (Study 2) reported to have heard more about medical algorithms than participants in the Prolific sample (Study 1), but the number of people who had tried a medical AI tool before was negligible in both samples (Study 1: Unfamiliar = 85\% [80/94], Not used = 13\% [12/94], Used = 2\% [2/94]; Study 2: Unfamiliar = 59\% [41/69], Not used = 38\% [26/69], Used = 3\% [2/69]). Interestingly, trust towards algorithms in general was comparable in the two samples, and, on average, participants in both samples had neutral attitudes (
measured on a scale of 0 to 10; Study 1: \textit{Mean} (\textit{M}) = 4.78, \textit{Standard Deviation} (\textit{SD}) = 2.18; Study 2: \textit{M} = 5.20, \textit{SD} = 2.00
). 
\BP{Since our sample sizes were defined by resource constraints, we conducted sensitivity power analyses as per the recommended good practice \cite{lakens2022sample}. The multilevel regressions directly comparing the SC with the OC or OT explanations (RQ2 and RQ3) were powered at the 80\% level to detect effect sizes of Cohen`s \textit{d} of 0.31 and 0.38, in Study 1 and 2, respectively.}\footnote{\BP{We used  GPower 3.1 for the sensitivity power analyses, and we calculated the effective sample sizes for each study by multiplying the number of participants by the cluster sizes (i.e. each user responded to two vignettes) and then dividing it by the design effect, which is 1+ (cluster size - 1) multiplied by the intra-class correlation (which we assumed to be low, r=0.2). Hence, we applied a design effect of 1.2 for our studies.}}




\paragraph{Attention Check}
We analysed the data from the most important feature task to assess whether the participants engaged with the experiment. 
Generally, the participants were able to correctly pinpoint the most important features in both studies. \BP{In Study 1, 72\%, whereas in Study 2, 86\% of the participants identified at least 3 out of the 4 most important features correctly. However, in Study 1, 5 participants failed to identify any of the 4 most important features, therefore, we did not include their data in any of the analyses.}

\paragraph{Objective Understanding of Explanations (RQ1)}
We merged the OC and OT explanations to a single \AR{Occlusion-1} explanation group, since format was not relevant for the definition recognition task. 
First, we ran two intercept only regression models to compare performance in recognising the SHAP and \AR{Occlusion-1} definitions against chance level. The \emph{OR} of recognising the correct SHAP definition remained around the chance level (\emph{OR} of 1/3) in both studies (Study 1: \emph{OR} = 0.46 [0.24, 0.76], Study 2: \emph{OR} = 0.32 [0.12, 0.87]). In contrast, the \emph{OR} of recognising the correct Occlusion-1 definition was substantially above the chance level in Study 2 (Study 1: \emph{OR} = 0.31 [0.10, 0.60], Study 2: \emph{OR} = 0.64 [0.38, 1.08]). 
We then analysed all data in a single regression model and used explanation type as a predictor. We found no evidence that participants were better at recognising 
\AR{Occlusion-1} over \AR{SHAP} explanations (Study 1: \emph{OR} = 1.14 [0.73, 1.78], \textit{p} = .566; Study 2: \emph{OR} = 1.27 [0.76, 2.13], \textit{p} = .359). 

\paragraph{Subjective Understanding of Explanations (RQ2)}
In both studies, the reported level of subjective understanding was higher for the \AR{OT} 
(Study 1: (\textit{M}) = 6.55, (\textit{SD}) = 2.39; Study 2: \textit{M} = 6.73, \textit{SD} = 2.21) than for the \AR{SC} explanations (Study 1: \textit{M} = 5.62, \textit{SD} = 2.51; Study 2: \textit{M} = 6.18, \textit{SD} = 2.62). In comparison, the level of reported understanding of the \AR{OC} explanations was comparable to that of the \AR{SC} explanations in Study 1 (\AR{OC} Study 1: \textit{M} = 5.49, \textit{SD} = 2.28), and higher than the reported understanding of \AR{SC} explanations in Study 2 (\AR{OC} Study 2: \textit{M} = 6.81, \textit{SD} = 2.13). The analyses revealed that when 
\AR{SC} 
explanations were compared to all \AR{Occlusion-1} explanations, participants perceived the \AR{latter} to be significantly more comprehensible than the \AR{former} in both studies. However, the comparison of the  different formats of the \AR{Occlusion-1} explanations to the \AR{SC} explanations in separate analyses only found evidence for the superiority of 
\AR{OT} over 
\AR{SC} explanations. Moreover, in Study 1, participants found 
\AR{OT} 
more comprehensible than 
\AR{OC} explanations. The 
other comparisons were non-significant. Table \ref{table:RQ2} presents the results of the multilevel regression analyses.

\begin{table}[t]
\centering
\begin{tabular}{ccccc}
\hline
\!\!Regression\!\! & 
\multicolumn{2}{c}{Study 1} & 
\multicolumn{2}{c}{Study 2} \\ 
Model & 
\textit{b} & 
\textit{p} &
\textit{b} & 
\textit{p} \\
\hline
\!\!\!\!\! \AR{S} v \AR{O} \!\!\!\!\! & 
\!\!\!0.45 [0.16, 0.73]\!\!\! &
\!\!\!\! .001 \!\! & 
\!\!\!0.58 [0.18, 0.98]\!\!\! &
\!\!\!\!\! .004\!\!\!\! \\
\!\!\AR{SC} v \AR{OC}\!\! & 
\!\!\!0.04 [-0.37, 0.45]\!\!\! &
\!\!\!\! .847 \!\! & 
\!\!\!0.51 [-0.07, 1.09]\!\!\! &
\!\!\!\!\! .088 \!\!\!\! \\
\!\!\AR{SC} v \AR{OT} \!\! & 
\!\!\!0.80 [0.42, 1.19]\!\!\! &
\!\!\!\! .001 \!\!  & 
\!\!\!0.67 [0.18, 1.16]\!\!\! &
\!\!\!\!\! .008 \!\!\!\! \\
\!\!\AR{OC} v \AR{OT} \!\! & 
\!\!\! 1.06 [0.19, 1.94] \!\!\! &
\!\!\!\! .030 \!\! & 
\!\!\! -0.09 [-0.97, 0.79] \!\!\! &
\!\!\!\!\! .841 \!\!\!\!\\
\hline
\end{tabular}
\caption{Multilevel regression models for the difference in subjective understanding of \underline{S}HAP and \underline{O}cclusion-1 explanations (\textbf{RQ2}). 
} 
\label{table:RQ2}
\end{table}


\paragraph{Trustworthiness of Explanations (RQ3)}
The reported level of trust was consistently the lowest for the \AR{SC} explanations across the studies (Study 1: \textit{M} \!=\! 4.82, \textit{SD} \!=\! 2.39; Study 2: \textit{M} \!=\! 6.73, \textit{SD} \!=\! 2.21). The \AR{OT} explanations scored the highest in Study 1 (Study 1: \textit{M} \!=\! 5.41, \textit{SD} \!=\! 2.33; Study 2: \textit{M} \!=\! 5.66, \textit{SD} \!=\! 2.18), and the \AR{OC} explanations scored the highest in Study 2 (Study 1: \textit{M} \!=\! 5.06, \textit{SD} \!=\! 2.24; Study 2: \textit{M} \!=\! 6.03, \textit{SD} \!=\! 1.97).
Similarly to the analyses of the subjective understanding of the explanations, when the \AR{Occlusion-1} explanations were compared 
with the \AR{SC} explanations, participants trusted the \AR{Occlusion-1} explanations significantly more in both studies. However, the follow-up analyses revealed some contradictions between the two studies. Namely, Study 1 showed evidence for the superiority of \AR{OT} over \AR{SC} explanations, while Study 2 showed evidence for the preference for \AR{OC} over \AR{SC} explanations in terms of their trustworthiness. The other comparisons remained non-significant. Table~\ref{table:RQ3} presents the results of the multilevel regression analyses.

\begin{table}[t]
\centering
\begin{tabular}{ccccc}
\hline
\!\!\!Regression\!\!\!\! & 
\multicolumn{2}{c}{Study 1} & 
\multicolumn{2}{c}{Study 2} \\ 
Model & 
\textit{b} &
\textit{p} &
\textit{b} &
\textit{p} \\
\hline
\!\!\AR{S} v \AR{O}\!\! & 
\!\!0.44 [0.17, 0.70]\!\! & 
\!\!\!\! .003 \!\! & 
\!\!0.38 [0.02, 0.75] \!\! & 
\!\!\!\!\! .042 \!\!\!\!\\
\!\!\AR{SC} v \AR{OC}\!\! & 
\!\!0.39 [-0.03, 0.80]\!\! & 
\!\!\!\! .066 \!\! & 
\!\!0.57 [0.04, 1.10] \!\! &
\!\!\!\!\! .035 \!\!\!\! \\
\!\! \AR{SC} v \AR{OT} \!\! & 
\!\!0.46 [0.12, 0.81] \!\! & 
\!\!\!\! .009 \!\! & 
\!\!0.23 [-0.25, 0.71] \!\! &
\!\!\!\!\! .345 \!\!\!\! \\
\!\!\AR{OC} v \AR{OT}\!\! & 
\!\!0.36 [-0.50, 1.22]\!\! &
\!\!\!\! .414 \!\! & 
\!\!0.37 [-0.48, 1.21] \!\!& 
\!\!\!\!\! .397\!\!\!\! \\
\hline
\end{tabular}
\caption{Multilevel regression models for the difference in the trustworthiness of \underline{S}HAP and \underline{O}cclusion-1 explanations \AR{(\textbf{RQ3})}. 
}
\label{table:RQ3}
\end{table}


\section{Discussion 
}
\label{sec:discussion}





Across two studies with different samples (from the general population and medical students), we found a reliable preference for \AR{Occlusion-1} over \AR{SHAP} explanations. Our participants reported that they understood and trusted the 
former more than the latter, as 
expected. However, the comparison of all the explanations revealed that this effect may be solely driven by the fact that our participants found the \AR{OT} explanations consistently more comprehensible than the \AR{SC} explanations. Moreover, in Study 1, the \AR{OT} 
explanations were also preferred to the \AR{SC}
explanations, though in Study 2 the results were less clear
. These findings are consistent with a notion of people, or more specifically patients within a healthcare setting, preferring explanations in text over chart format \cite{Bertrand_23}, for instance, due to finding some graphs difficult to interpret \cite{durand2020graph,
Szymanski_21}. 
These findings indicate that prioritising FA explanations based on  simple rather than complex models may not contribute to clarity. 
Interestingly, the results regarding trustworthiness were slightly in conflict with this idea, since in our medical student sample we only found evidence for higher trust in the \AR{OC} \BP{compared to the SC explanations. }
While these findings show support for the \BP{role of explanation content}
, more research is needed to explore how people relate to and utilise different XAI techniques, and how \BP{and when} an explanation's format 
\BP{
trumps the impact of its content.} 
Future research should also explore the potential of tailoring explanation \BP{content} and (possibly different types of) format to individuals \cite{Lesley_24}, e.g. based on their expertise in a field \cite{Wang_23} 
and their skills in interpreting numerical information (e.g. graph literacy  or numeracy \cite{
durand2020graph}
). 

Our studies did not assess SHAP explanations in text format, since, as we discussed, it is not obvious how to create an ``ST'' explanation that is representative of SHAP and comprehensible.
In any case, even if we could have included these additional explanations, the relatively lower sample size of Study 2 may have caused some of our tests to be insensitive and non-significant, preventing us from settling the question of whether \AR{Occlusion-1} explanations are perceived to be more comprehensible than \BP{SHAP} explanations regardless of the format, 
e.g. challenging the idea that the preference for \AR{Occlusion-1} explanations is driven by a preference for charts. 

\BP{Our expert sample in Study 2 comprised medical students 
so it is not obvious how well our findings generalise to experienced healthcare practitioners
}: we leave this to future work. Also, self-report measures are prone to biases: Nagendran et al.~\cite{nagendran2023quantifying} found no correlation between self-reported understanding of XAI and the impact of XAI on behaviour, casting some doubt on their generalisability to real-life behaviour. Future research should investigate this presumed dissociation for behavioural measures of advice understanding and uptake \cite{palfi2022algorithm,Panigutti_22}. 
This justifies our definition of a novel performance measure 
  (i.e.~the definition recognition task
  ), especially since we found no evidence that participants were better at recognising the definition of either explanation: it would be interesting to study variants of this measure or different ways to assess it to \BP{further explore the impact of explanation content.}
 The task of choosing the correct definition of explanations may have been too difficult as performance was only reliably over chance level for the \AR{Occlusion-1} explanations in the medical student sample, potentially creating a floor effect. 
 \BP{Given the complexity of the definitions and their technical language, failing to identify the correct definition is not necessarily a strong indicator of a lack of understanding of the explanations.}
 \ar{Further, it would be interesting to study the effect of users' awareness of XAI methods' vulnerability to manipulation and capability for producing misleading explanations.}

In summary, by evaluating 
explanations across various realistic scenarios, metrics, and user groups, our 
approach makes a novel contribution to the understanding of how XAI techniques can be 
best used in healthcare. 
While our experiments did not provide 
definitive evidence for the dominance of Occlusion-1 over SHAP explanations, 
this lack of a clear preference is interesting in itself, especially 
as the simpler Occlusion-1 explanations were expected to be favoured. These findings underscore the importance of further research and empirical evaluation on the interplay between explanation content and format, pointing to several future directions.




\section*{Acknowledgments}

Rago and Toni were partially funded by the ERC under the European
Union’s Horizon 2020 research and innovation programme (grant no. 101020934)
{and by J.P. Morgan and the Royal Academy
of Engineering under the Research Chairs and Senior Research
Fellowships scheme (grant no. RCSRF2021\textbackslash 11\textbackslash 45)}.




\bibliography{bib_short}

\newpage

\section*{Supplementary Material}

Following are three examples of other vignettes used in the experiments.

\begin{example}
\label{ex:app1}

    Patient name: Rish Barrett (male),
    Age: 60,
    Current BMI: 37.55 (Height 175cm, Weight 115kg),
    Smoking: Smoker 5 cigarettes/day, 
    Past medical history: Type 2 Diabetes Mellitus, 
    Family history: Nil relevant.

    Mr Barrett comes to see you for his results. He saw another GP at the surgery two weeks ago about the fact that after a long time he had lost a lot of weight within a short period of time (2 months) and could not understand why. Your colleague had organised some investigations. During your consultation, he reports that he has ongoing weight loss and has also been experiencing some abdominal pain over the past month. There are no other symptoms and examination findings are normal.

    The cancer prediction algorithm estimates the risk of gastro-oesophageal cancer to be 2.53\%.
    
\end{example}

\begin{example}
\label{ex:app2}

    Patient name: Natalie Clark (female),
    Age: 74,
    Current BMI: 34.60 (Height 170cm, Weight 100kg),
    Smoking: Ex-smoker,
    Past medical history: Nil,
    Family history: Nil relevant.
    
    Natalie sees you today accompanied by her husband. Over the past 6 weeks, she has been experiencing heartburn and has been off her food. She has vomited on a few occasions and seen a bit of blood in the vomit. She also finds it difficult to swallow when eating. She has no other symptoms and examination findings are normal.

    The cancer prediction algorithm estimates the risk of gastro-oesophageal cancer to be 40.24\%.

\end{example}

\begin{example}
\label{ex:app3}

    Patient name: Marianne Foster (female),
    Age: 60,
    Current BMI: 27.85 (Height 163cm, Weight 74kg),
    Smoking: Never smoked,
    Past medical history: Nil,
    Family history: Nil relevant.

    Mary comes back to see you for a follow-up consultation. Another GP at the surgery had organised some blood tests and a FIT test 4 weeks ago. On the electronic health record, you can see that during the last consultation she had complained of having lost her appetite as well as some weight (about 3kg over the previous 6 weeks) despite no changes in her diet or lifestyle. She reports that she is still off her food and has lost more weight. She has also noticed that it is becoming increasingly difficult for her to swallow when she tries to eat. The blood tests and FIT test which were organised came back normal. She denies any post-menopausal bleeding, has no other symptoms and examination findings are normal.

    The cancer prediction algorithm estimates the risk of gastro-oesophageal cancer to be 8.27\%.
    
\end{example}

Table \ref{table:qcancer_app} summarises the values they result in for the QCancer algorithm.


\begin{table*}[b]
\centering
\begin{tabular}{cccccccccccccccc}
\cline{2-16}
 &
Gender & 
Age & 
BMI &
SS &
T2 &
CP &
LA &
WL &
DS &
AP &
BV &
I &
H &
A & 
BH \\ 
\hline
Ex. \ref{ex:app1} &
M &
60 &
37.55 &
light &
yes &
no &
no &
yes &
no &
yes &
no &
no &
no &
no &
no \\
Ex. \ref{ex:app2} &
F &
74 &
34.60 &
ex &
no &
no &
yes &
no &
yes &
no &
yes &
no &
yes &
no &
no \\
Ex. \ref{ex:app3} &
F &
60 &
27.85 &
non &
no &
no &
yes &
yes &
yes &
no &
no &
no &
no &
no &
no \\
\hline
\end{tabular}
\caption{Values used for patients described in Examples \ref{ex:app1}-\ref{ex:app3} (See Table \ref{table:qcancer} for the variables).}
\label{table:qcancer_app}
\end{table*}

\begin{figure}
    In the following, we give an example of the survey which was presented to users. First, users were asked for their consent:

    \centering
    \includegraphics[width=1\linewidth]{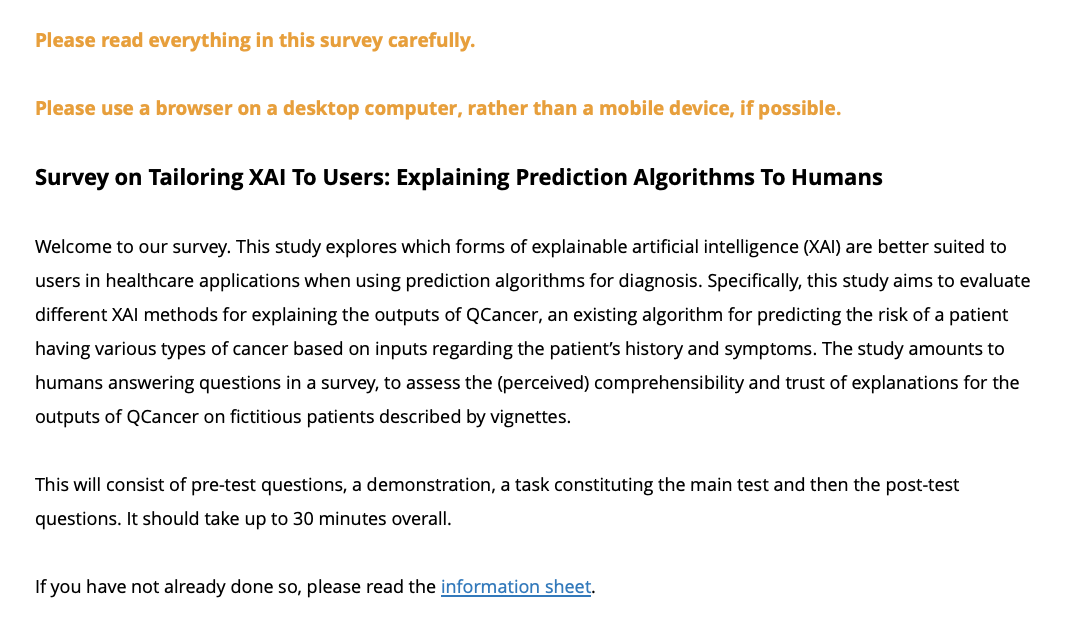}
    \includegraphics[width=1\linewidth]{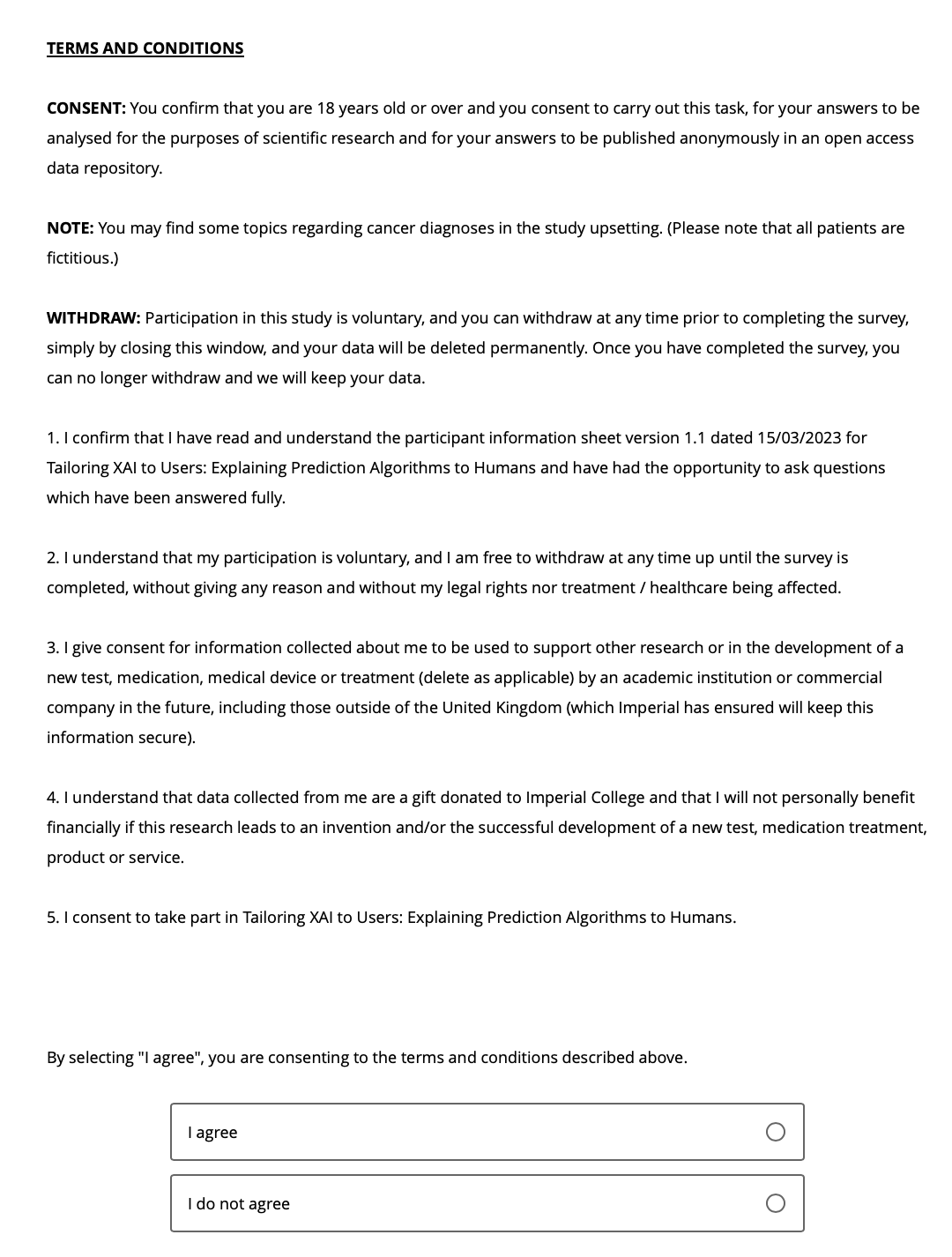}
\end{figure}


\raggedright

\begin{figure}
    We then asked the users the pre-test questions:

    \centering
    \includegraphics[width=0.9\linewidth]{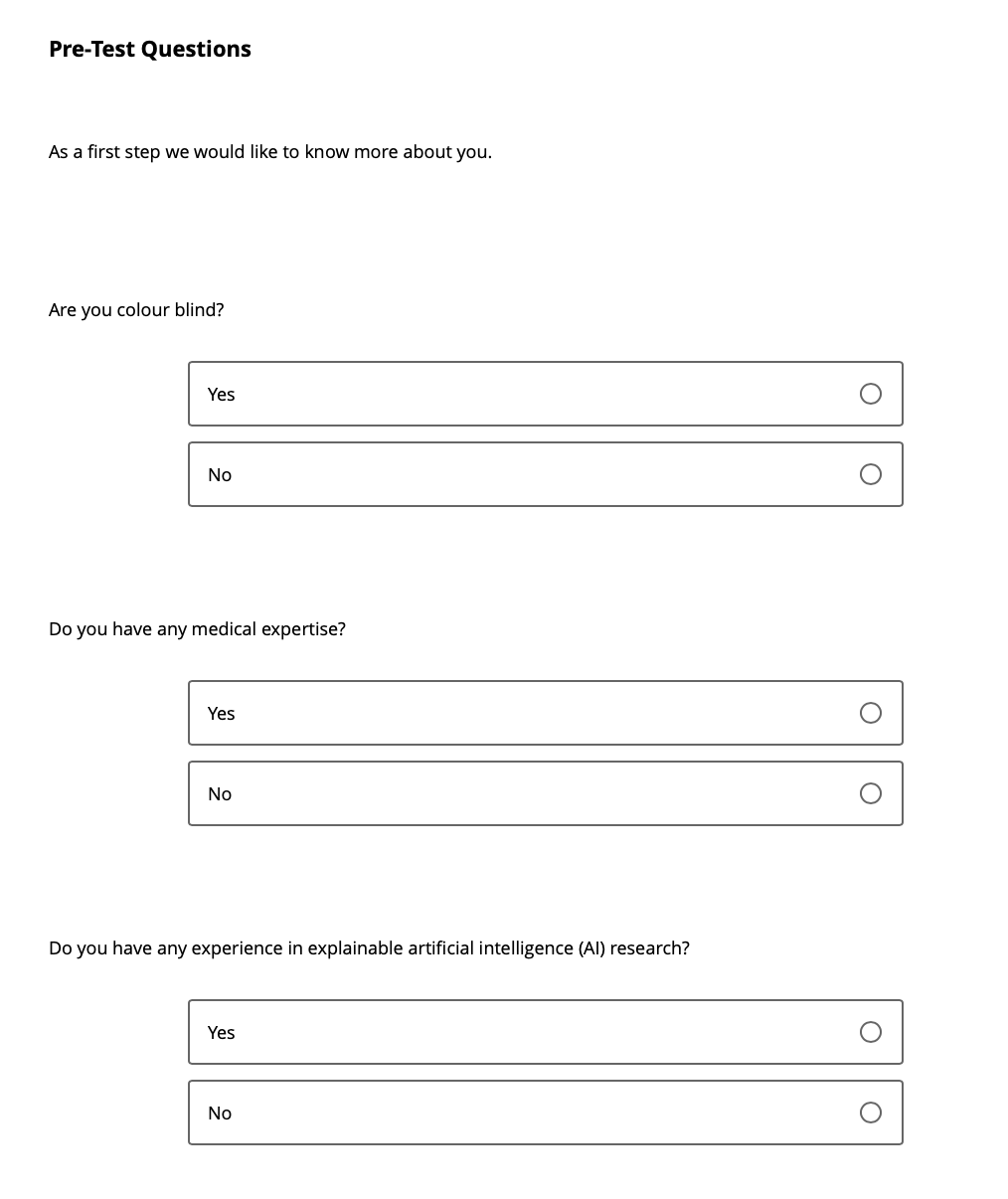}
    \includegraphics[width=1\linewidth]{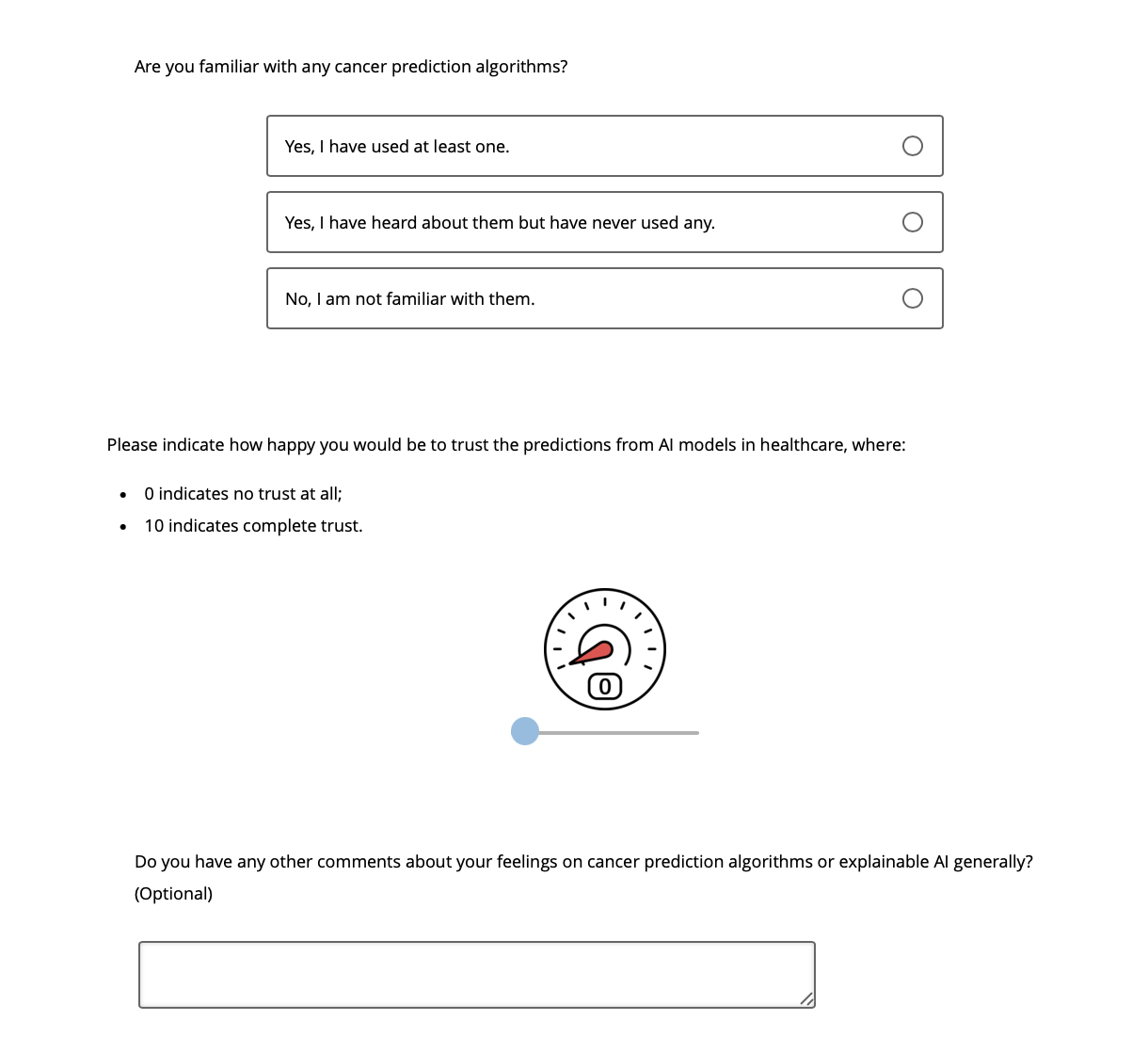}
\end{figure}


\raggedright

\begin{figure}
    We then showed the users a demonstration of the main test questions:
    
    \centering
    \includegraphics[width=1\linewidth]{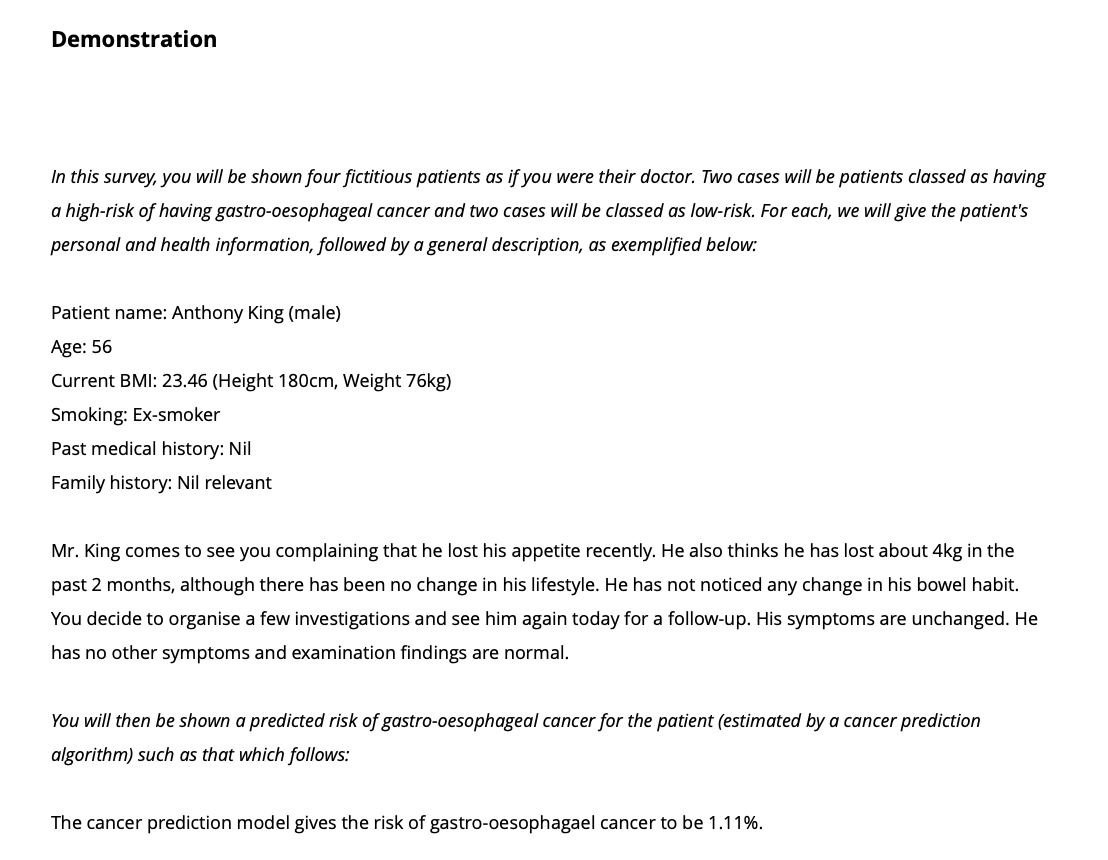}
    \includegraphics[width=1\linewidth]{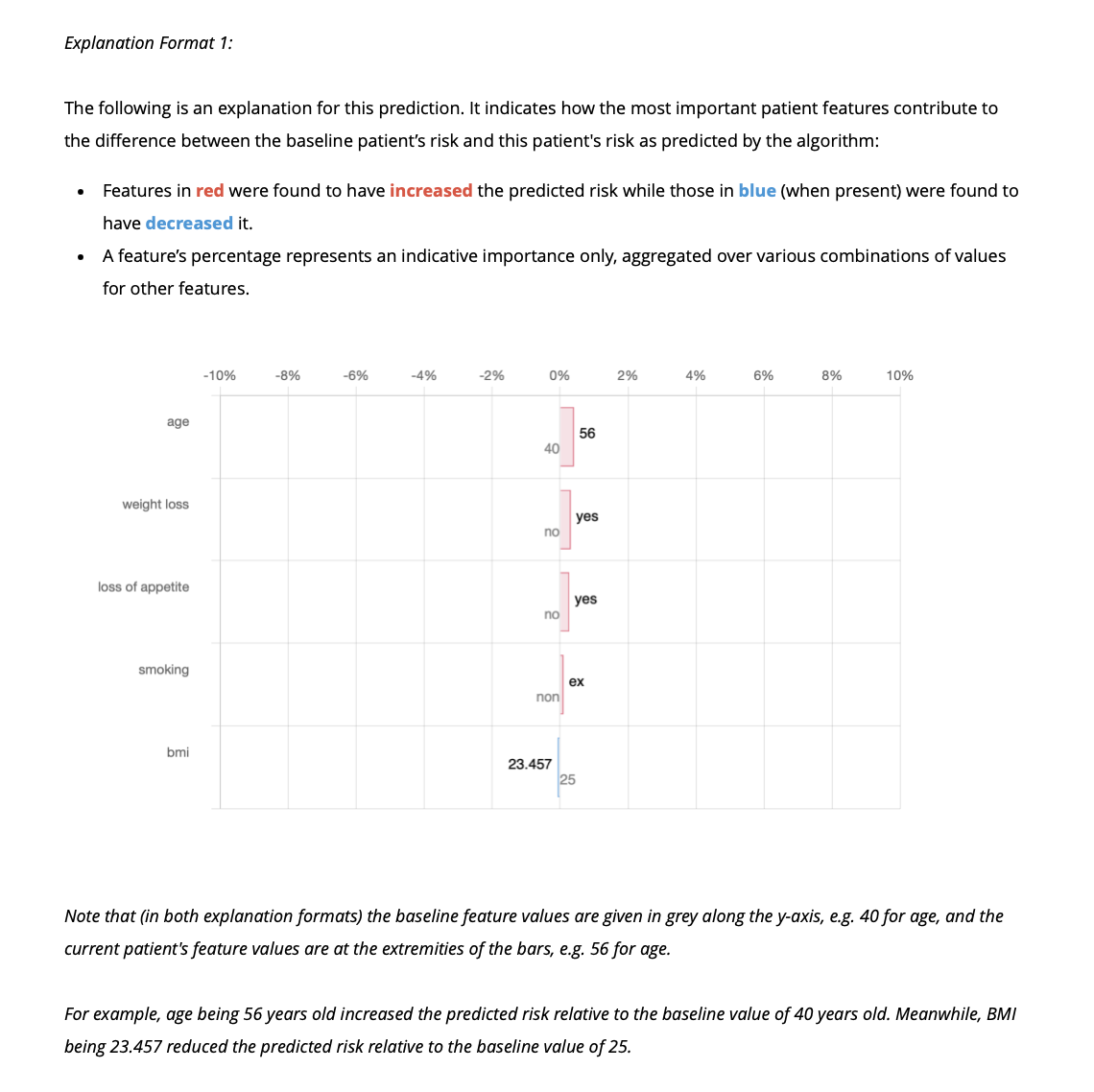}


    \includegraphics[width=1\linewidth]{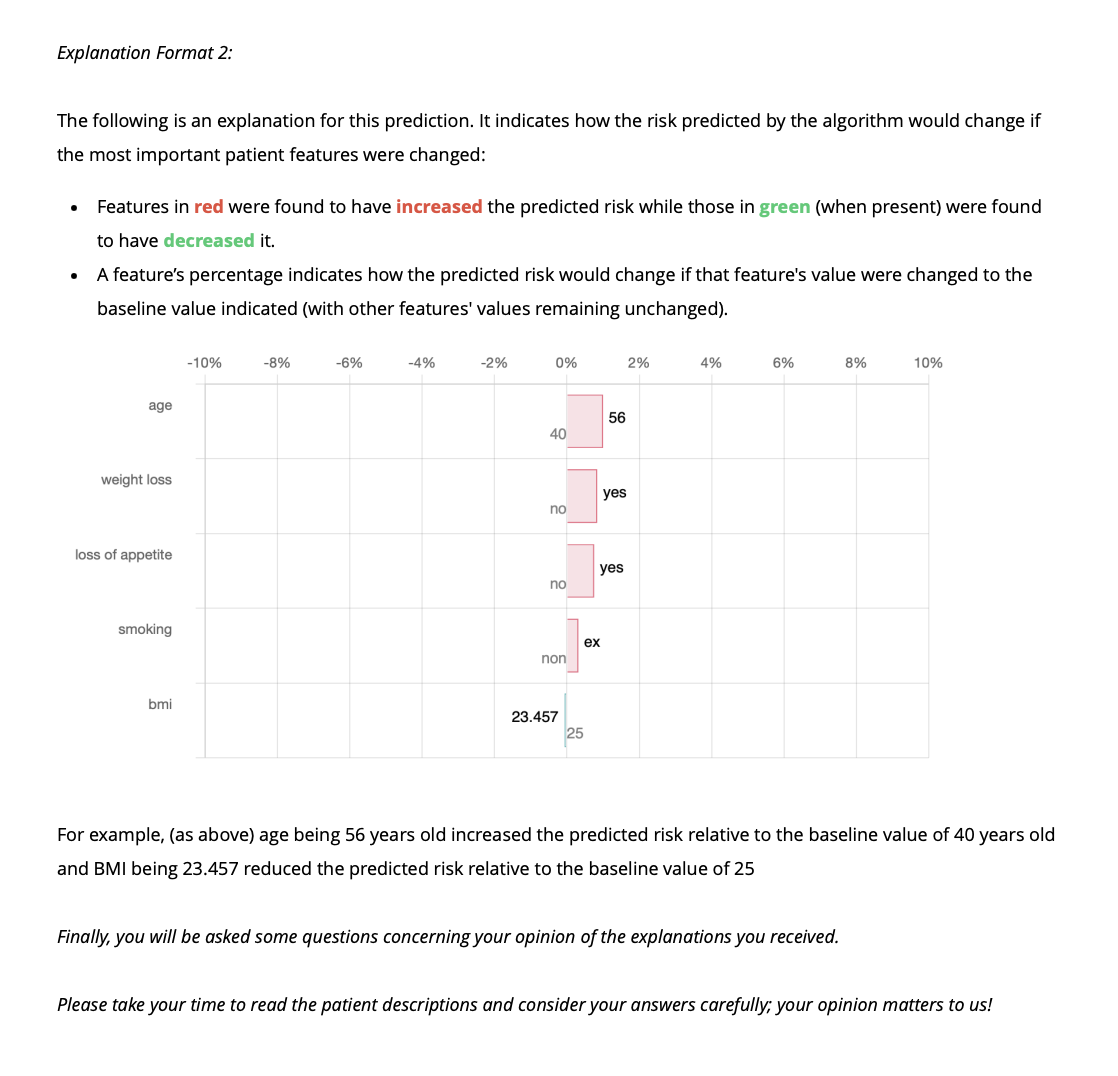}
\end{figure}

\newpage

\raggedright

\begin{figure}
    For each set of vignette, prediction and explanation in the main test (which were as seen in the demonstration), we asked the following questions, with the last being on a new page.
    
    \centering
    \includegraphics[width=1\linewidth]{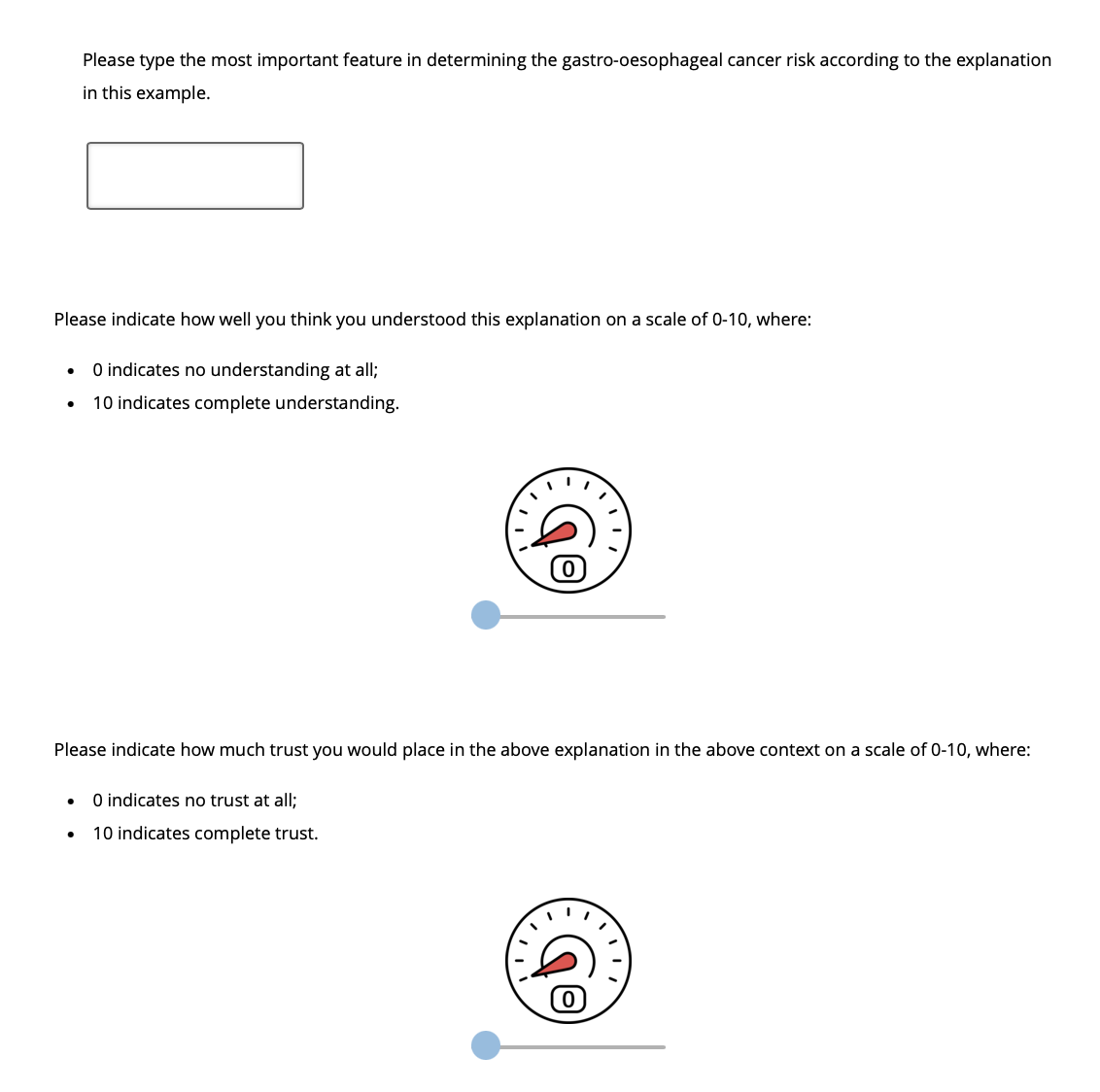}
    \includegraphics[width=1\linewidth]{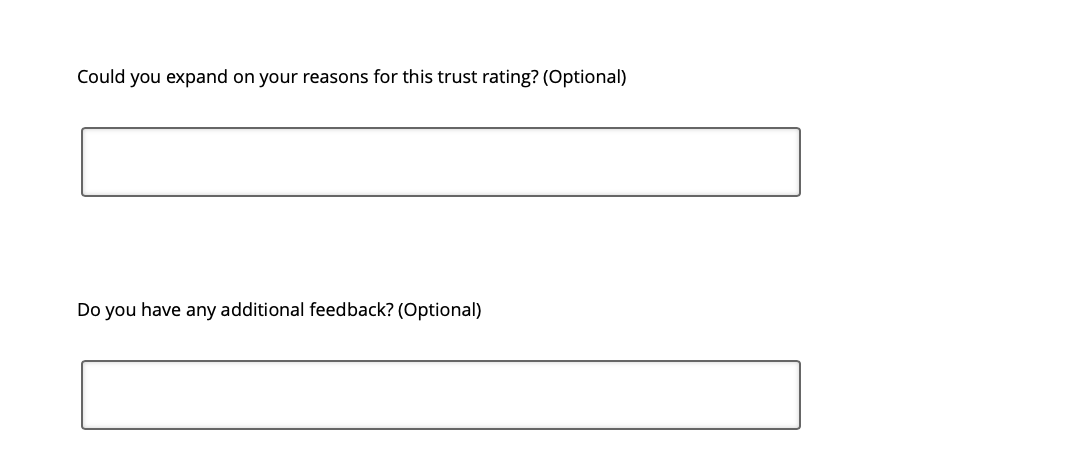}
    \includegraphics[width=1\linewidth]{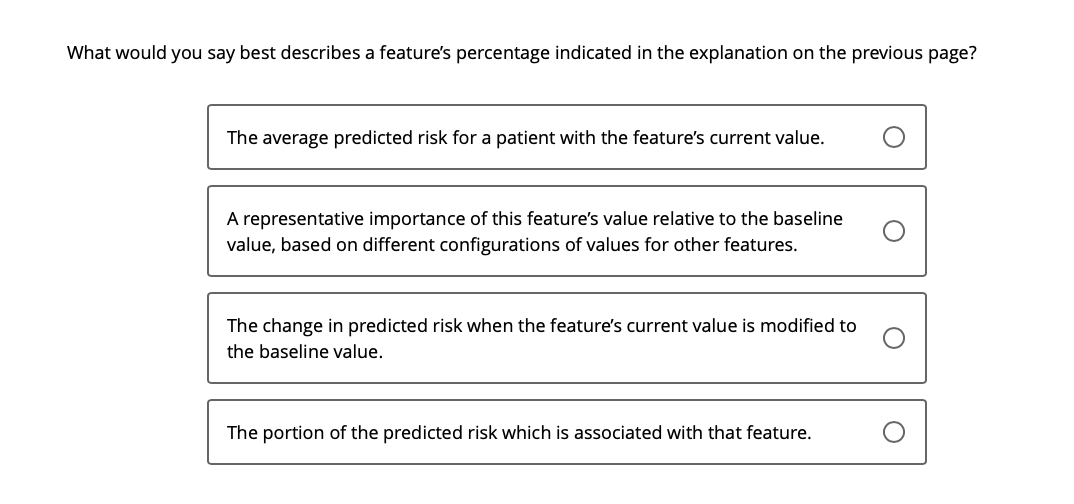}
\end{figure}

\newpage

\raggedright

\begin{figure}
    Finally, we asked users the post-test questions:
    
    \centering
    \includegraphics[width=1\linewidth]{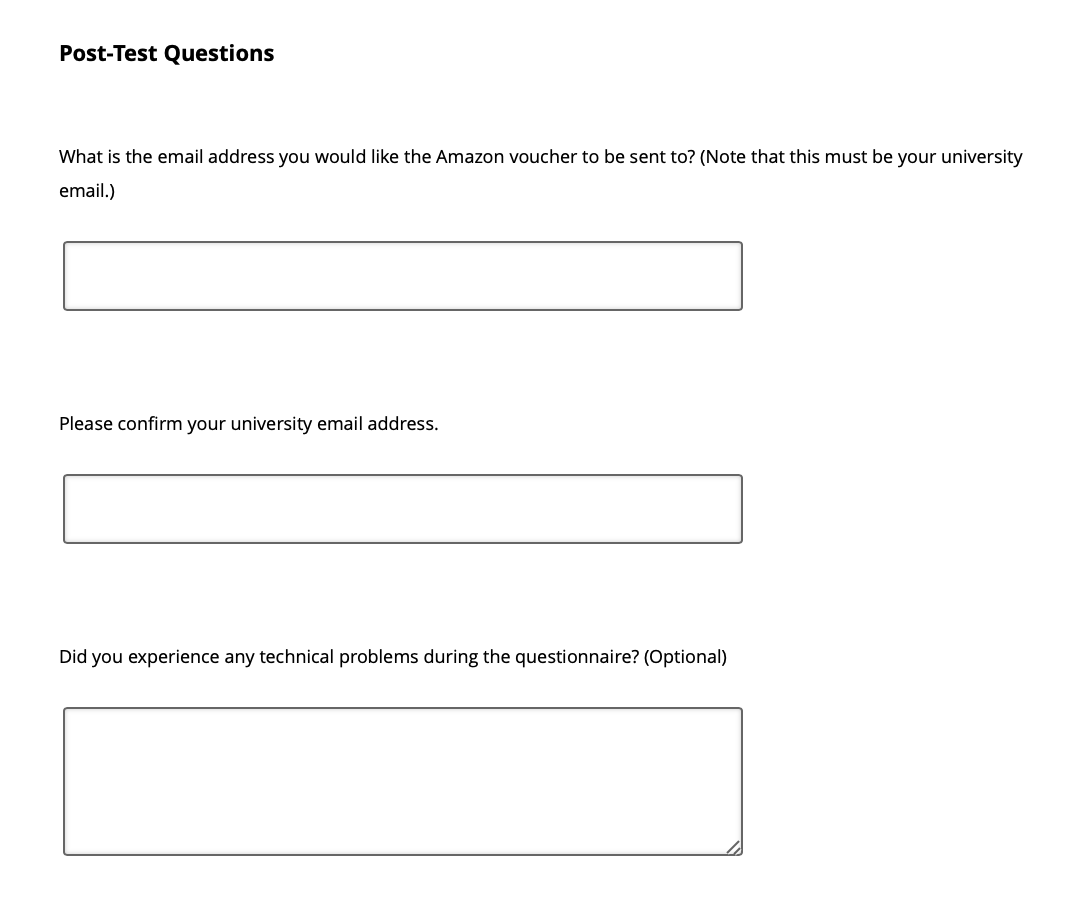}
    \includegraphics[width=1\linewidth]{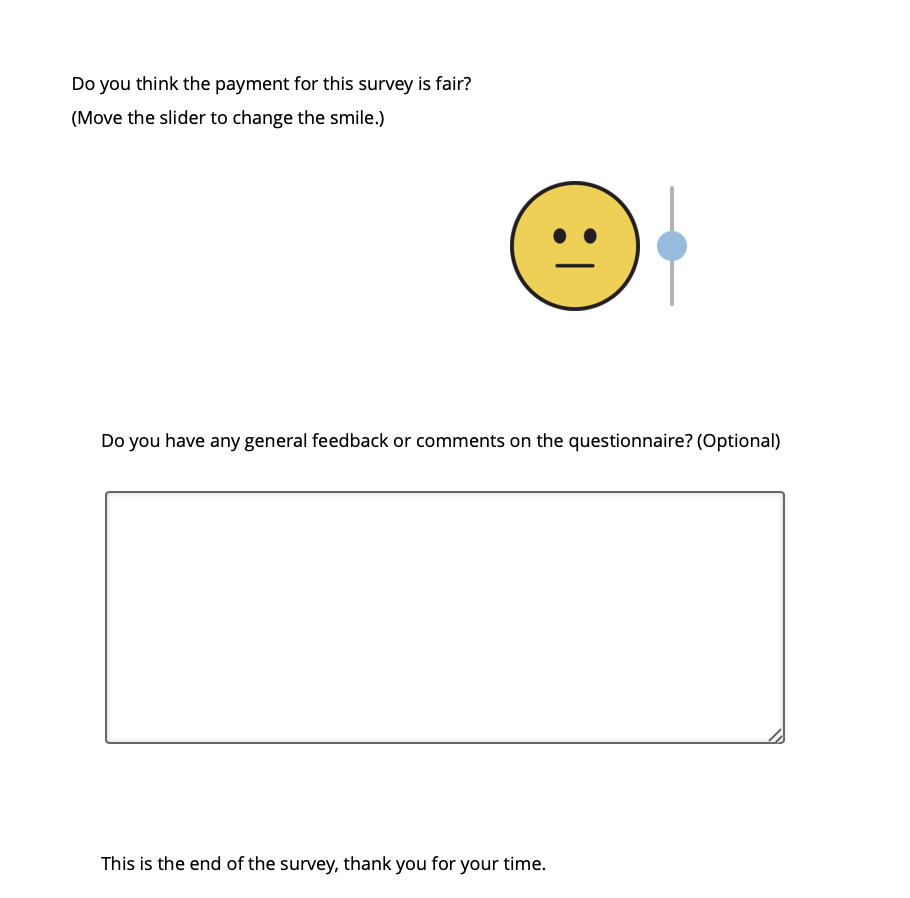}
\end{figure}

\raggedright










\end{document}